\documentclass[final,5p,times,twocolumn]{elsarticle}

\usepackage{amssymb}
\usepackage{amsmath}
\usepackage{latexsym}
\usepackage{acronym}
\usepackage{multirow}
\usepackage{url}

\usepackage{booktabs}
\usepackage{comment}
\usepackage{bbm}
\usepackage{amssymb}
\usepackage{dsfont}
\usepackage{stfloats}
\usepackage{graphicx}
\usepackage{tabularx}
\usepackage{algorithm}
\usepackage{algpseudocode}

\usepackage[table]{xcolor}
\usepackage{xcolor}
\definecolor{ours}{RGB}{224,224,255}

\acrodef{AL}[AL]{Active Learning}
\acrodef{BiLSTM}[LSTM]{Bidirectional Long Short-Term Memory}
\acrodef{CNN}[CNN]{Convolutional Neural Network}
\acrodef{DDS}[DDS]{Denoised Distant Dataset}
\acrodef{DOREMI}[DOREMI]{\textbf{DO}cument-level \textbf{R}elation \textbf{E}xtraction opti\textbf{M}izing the long ta\textbf{I}l}
\acrodef{DDS}[DDS]{Denoised Distantly Supervised Dataset}
\acrodef{DS}[DS]{Distant Supervision}
\acrodef{DocRE}[DocRE]{Document-Level Relation Extraction}
\acrodef{ER}[ER]{Evidence Retrieval}
\acrodef{GCN}[GCN]{Graph Convolutional Network}
\acrodef{GNN}[GNN]{Graph Neural Network}
\acrodef{GAT}[GAT]{Graph Attention Network}
\acrodef{IE}[IE]{Information Extraction}
\acrodef{KBC}[KBC]{Knowledge Base Construction}
\acrodef{KG}[KG]{Knowledge Graph}
\acrodef{LLM}[LLM]{Large Language Model}
\acrodef{LSTM}[LSTM]{Long Short-Term Memory}
\acrodef{NER}[NER]{Named Entity Recognition}
\acrodef{NER+L}[NER+L]{Named Entity Recognition and Linking}
\acrodef{NLI}[NLI]{Natural Language Inference}
\acrodef{NLP}[NLP]{Natural Language Processing}
\acrodef{PPD}[PPD]{Positive Probability Disagreement}
\acrodef{PPM}[PPM]{Positive Probability Mean}
\acrodef{RE}[RE]{Relation Extraction}
\acrodef{SC}[SC]{Selection Criteria}

\journal{Pre-print version of authors}

\begin{document}

\begin{frontmatter}



\title{DOREMI: Optimizing Long Tail Predictions in Document-Level Relation Extraction\tnoteref{*}}
\tnotetext[*]{\textbf{Paper accepted for publication in Knowledge-Based Systems.}}


\author[label1]{Laura Menotti\corref{cor1}} 
\ead{laura.menotti@unipd.it}
\author[label1]{Stefano Marchesin}
\ead{stefano.marchesin@unipd.it}
\author[label1]{Gianmaria Silvello}
\ead{gianmaria.silvello@unipd.it}

\affiliation[label1]{organization={Department of Information Engineering},
            addressline={University of Padova}, 
            city={Padova},
            country={Italy}}

\cortext[cor1]{Corresponding author.}

\begin{abstract}
\ac{DocRE} presents significant challenges due to its reliance on cross-sentence context and the long-tail distribution of relation types, where many relations have scarce training examples. In this work, we introduce \ac{DOREMI}, an iterative framework that enhances underrepresented relations through minimal yet targeted manual annotations. Unlike previous approaches that rely on large-scale noisy data or heuristic denoising, \ac{DOREMI} actively selects the most informative examples to improve training efficiency and robustness. \ac{DOREMI} can be applied to any existing \ac{DocRE} model and is effective at mitigating long-tail biases, offering a scalable solution to improve generalization on rare relations.
\end{abstract}




\end{frontmatter}



\section{Introduction}
\begin{figure}[t]
    \centering
    \includegraphics[width=\columnwidth]{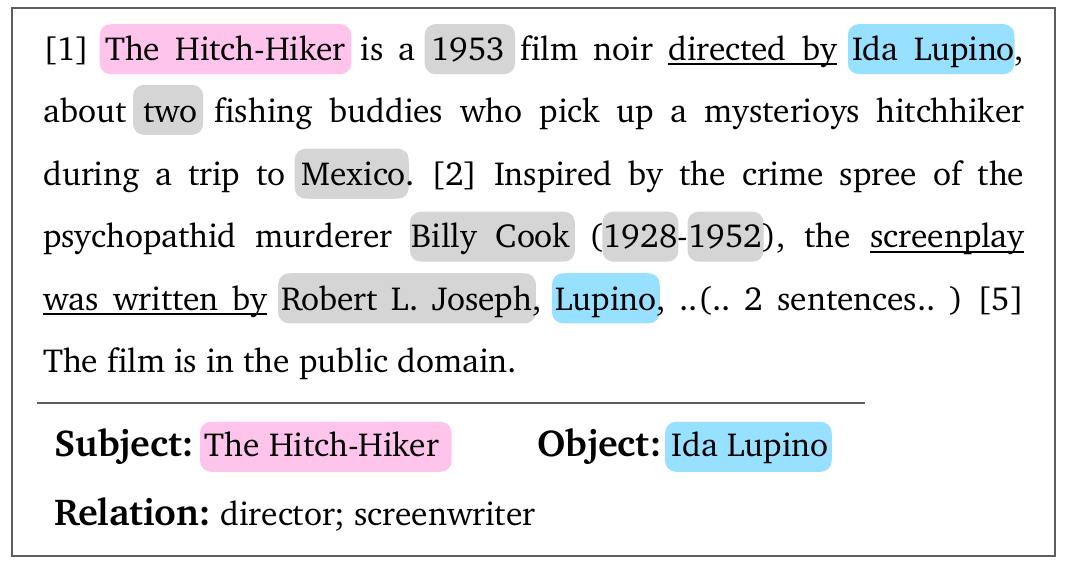}
    \caption{An example of multi-entity and multi-label problems from the DocRED dataset. Subject entity \emph{"The Hitch-Hiker"} (in pink) and object entity \emph{"Ida Lupino"} (in blue) express two relations in the document (\emph{director} and \emph{screenwriter}). Other entities are in grey.}
    \label{fig:example}
\end{figure}

\acf{DocRE} is an NLP task that identifies all relations between a given entity pair within a document. \ac{DocRE} is a more realistic setting than the more commonly studied \emph{sentence-level} \ac{RE}, as many relational facts are typically expressed across multiple sentences. 
Sentence-level \ac{RE} considers only a single entity pair connected by a maximum of one relation; whereas, \ac{DocRE} involves documents that contain multiple entity pairs, each of which can appear multiple times and be associated with different relations. 
Figure~\ref{fig:example} shows the typical case where an entity pair has multiple relations spanning multiple sentences; indeed, the pair (\emph{The Hitch-Hiker}, \emph{Ida Lupino}) has two relations: ``\emph{director}" derived from the first sentence and ``\emph{screenwriter}" inferred from combining the first and the second sentence. This case is common, for instance, around 40\% of relational facts in Wikipedia documents can only be extracted from multiple sentences~\cite{yao_etal-2019}. 

The reference test collection for the task is DocRED, a large-scale dataset constructed from Wikipedia and Wikidata~\cite{yao_etal-2019}. It comprises two training datasets: one includes 3,053 manually annotated documents, while the other $101,873$ documents are annotated using \ac{DS}. Although DocRED is commonly used for evaluating \ac{DocRE} models, recent works showed that the collection struggles with false negatives~\cite{huang_etal-2022, tan_etal-2022b}. 
\citet{tan_etal-2022b} found out that almost 65\% of ground truth relations were not annotated in DocRED; they corrected the manual training and development set and released Re-DocRED, meant as an improved version of DocRED.

\begin{figure*}[t]
    \centering
    \includegraphics[width=\textwidth]{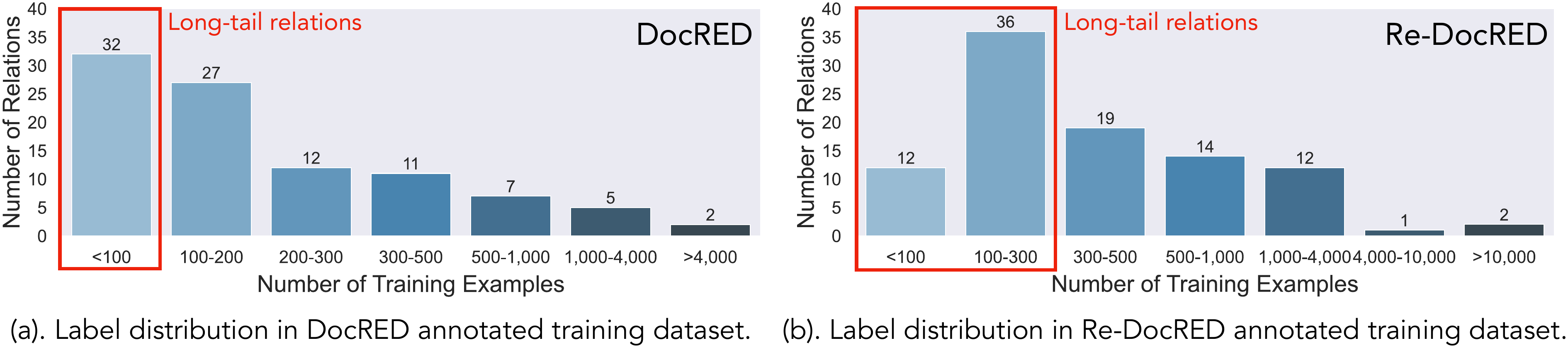}
    \caption{Label distribution in DocRED (a) and Re-DocRED (b) annotated training dataset.} 
    \label{fig:lbl-distribution}
\end{figure*}
Nevertheless, DocRED and Re-DocRED show a significantly imbalanced distribution of training examples across the 96 annotated relations~\cite{han_etal-2024}.
Figure~\ref{fig:lbl-distribution} shows the distribution of labels in the annotated training datasets of both reference datasets. They exhibit a typical power-law pattern, where a small number of relations -- i.e., \emph{frequent relations} -- have many training examples, while the majority have only a few -- i.e., \emph{long-tail relations}. The DocRED annotated training dataset consists of $38,180$ examples, half representing the four most frequent relations, while $31$ relations (out of $96$) have fewer than $100$ training examples each. This class imbalance is even more pronounced in Re-DocRED, where the training dataset contains $85,932$ examples (three times more than DocRED), with $20,402$ ($24\%$ of all instances) about a single relation. In addition, half of the relations ($48$ out of $96$) have fewer than $300$ examples each, of which 12 (almost 13\% of all relations) have fewer than $100$.
This class imbalance raises concerns about the actual effectiveness of the models, whose performance is biased toward frequent relations, at the expense of accurately identifying long-tail relations~\cite{han_etal-2024,choiEtAl2023}. 

One potential approach to improve models' performance on the long-tail is to use \ac{DS} data to augment the number of training examples for underepresented relations. However, the DocRED \ac{DS} dataset is noisy and biased toward popular relations. This dataset was created by aligning Wikipedia documents with Wikidata triples, where the inclusion of entities and properties tends to favor frequently mentioned concepts~\cite{huang_etal-2022}. 
The most effective label denoising strategy in \ac{DocRE} is UGDRE~\cite{sun_etal-2023}, leveraging uncertainty accuracy estimation based on Monte Carlo dropout. However, it does not consider long-tail relations, requiring strategies that can selectively enhance underrepresented relations. 
Although promising, \acp{LLM} still underperform compared to current state-of-the-art approaches for \ac{DocRE}, making them ineffective for the task~\cite{zhang_etal-2025a}.
The current best performing approaches for \ac{DocRE} features ATLOP~\cite{zhou_etal-2021}, a sequence-based model introducing an adaptive thresholding and localized context pooling, and DREEAM~\cite{ma_etal-2023}, enhancing predictions with an evidence-based attention network in a teacher-student paradigm.

\begin{figure*}[t]
    \centering
    \includegraphics[width=\textwidth]{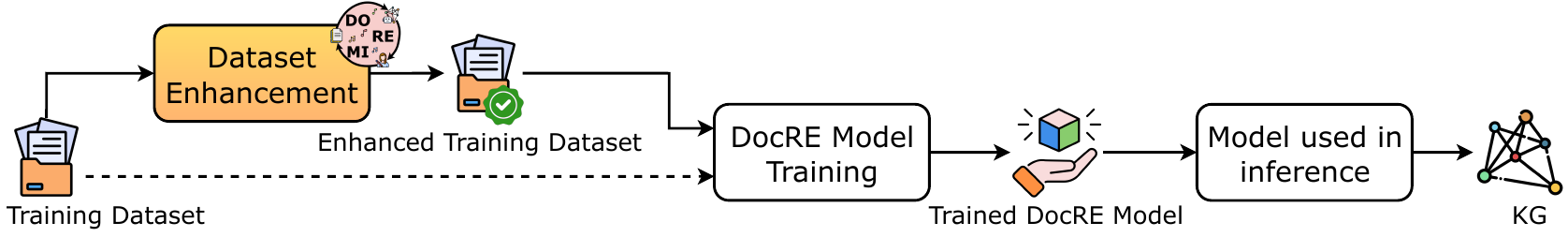}
    \caption{The general pipeline of a DocRE model. DOREMI enhances the training dataset, optimizing long-tail relations. Once the enhanced dataset is obtained, it can be used for training any DocRE model.}
    \label{fig:scope}
\end{figure*}

This work proposes \acf{DOREMI}, an iterative system that enhances the training data through targeted manual annotation of highly informative examples selected leveraging a disagreement-based criteria inspired by~\cite{le_etal-2024}. DOREMI differs from prior denoising strategies for its iterative refinement architecture and the disagreement-based instance selection for manual annotation. To our knowledge, this marks the first effort to tailor an active learning approach specifically for \ac{DocRE}, featuring a human-in-the-loop strategy for denoising. 
As shown in Figure~\ref{fig:scope}, \ac{DOREMI} operates upstream in the general DocRE pipeline by augmenting the dataset in a model-agnostic fashion, enabling any downstream \ac{DocRE} model to benefit from improved long-tail coverage. To ensure high-reliability extraction, \ac{DOREMI} is precision-oriented in its design. In real-world applications of automatic \ac{KBC}, prioritizing precision over recall is crucial to ensure the reliability of extracted relations. Including fewer but more accurate triples minimizes the risk of introducing incorrect information, which can significantly degrade downstream tasks~\cite{dalvi_etal-2017}.
Experimental results on the DocRED development dataset indicate that annotating merely \emph{0.001\%} of the \ac{DS} dataset in DocRED (400 triples in total) significantly enhances the performance of leading models like ATLOP and DREEAM. Compared to the leading label denoising technique, that is, UGDRE, \ac{DOREMI} registers an increase in overall \textsc{precision} and \textsc{recall} of up to $+8.2\%$ and $+3.3\%$, respectively (\textsc{F1} improves by up to $+0.7\%$). For long-tail triples ($<100$ examples in training), \textsc{precision} increases by $+76.0\%$ and \textsc{F1} by $+5.0\%$. In long-tail triples involving entity pairs not seen in training, \ac{DOREMI} boosts \textsc{ignF1} by up to $28.7\%$ and \textsc{ignPrecision} by $137.6\%$ with respect to UGDRE.
The improvements are confirmed on the Re-DocRED test dataset, with a gain of \textsc{precision} and \textsc{ignPrecision} up to $+16.2\%$ and $+19.2\%$ on long-tail triples ($<300$ examples in training) from \emph{annotating only 0.003\%} of the distant dataset ($1,200$ triples). Moreover, if we consider \emph{extreme long-tail triples} in Re-DocRED ($<100$ examples in training, \ac{DOREMI} registers a substantial improvement in \textsc{precision} ($+83.2\%$), \textsc{ignPrecision} ($+207.9\%$), \textsc{F1} ($1.3\%$), and \textsc{ignF1} ($+33.5\%$). 
The hybrid setting, i.e., considering UGDRE predicitions for frequent relations and \ac{DOREMI} for the long-tail, registers an improvement in overall \textsc{precision} and \textsc{ignPrecision} of up to $+3.7\%$ and $+4.5\%$, respectively. These results indicates \ac{DOREMI} effectively complements exisiting denoising approaches.

The core \textbf{contributions} of this work are listed as follows: (1) We propose \ac{DOREMI}, a novel iterative system tailored for long-tail relations to enhance the distantly supervised dataset through disagreement-driven annotations (Section~\ref{sec:methodology}); (2) 
We demonstrate for the first time that measuring the disagreement between multiple models is a good proxy to identify Hard-To-Classify examples specifically for \ac{DocRE} and yields substantial performance improvements with negligible human effort (Section~\ref{subsec:sampling}); (3) We release two new \acp{DDS}, one based on DocRED and one on Re-DocRED, which can be used to train any DocRE model~\cite{doremi_zenodo}. Unlike existing resources, these datasets are explicitly optimized for long-tail relations and can be directly used to train arbitrary DocRE models, leading to consistent and significant improvements as validated by extensive experiments.(Section~\ref{sec:results}). The implementation of \ac{DOREMI} is available in GitHub at: \url{https://github.com/mntlra/DOREMI}.

The rest of this work is organized as follows. Section~\ref{sec:related} presents previous efforts in \ac{DocRE}, with special attention to label denoising and sequence-based models. Section~\ref{sec:methodology} describes \ac{DOREMI}.
Section~\ref{sec:results} reports the performance of \ac{DOREMI} compared to other denoising strategies evaluated on the DocRED development dataset (Section~\ref{subsec:docred}) and Re-DocRED test dataset (Section~\ref{subsec:redocred}). Section~\ref{sec:discussion} reports an ablation study about disagreement-based sampling and some statistics on the manual annotation process. To conclude, section~\ref{sec:conclusion} draws some final remarks.

\section{Related Work}
\label{sec:related}
\ac{DocRE} models are categorized into graph-based and sequence-based approaches.  

\textbf{Graph-based methods} construct a document graph where nodes represent words, mentions, entities, or sentences, and edges capture various dependencies. Early approaches used syntactic dependencies to build these graphs~\cite{quirk_poon-2017}. \citet{christopoulou_etal-2019} introduced the edge-oriented graph model, which employs different types of edges to capture intra- and inter-sentential relations. Subsequent models, like GAIN~\cite{zeng_etal-2020}, constructed both mention-level and entity-level graphs to perform multi-hop reasoning for relation extraction. These methods leverage graph neural networks to propagate information across the graph and infer relations between entities. In this context, SSAN introduces self-attention to incorporate the coreference and co-occurrence structure of entities into training~\cite{xu_etal-2021a}. DocuNet instead considers \ac{DocRE} as a semantic segmentation task and exploits an entity-level relation matrix to capture local and global information~\cite{zhang_etal-2021}. \citet{xu_etal-2021b} proposes a reconstruction mechanism to allow the model to capture path dependencies between an entity pair and its ground-truth relationship. \citet{wang_etal-2024} leverages Hierarchical Dependency Tree and Bridge Path (HDT-BP) to represent fine-grained features aiding relation predictions. DocRE-CLiP frames \ac{DocRE} as a link prediction problem combining link prediction with contextual knowledge and achieves state-of-the-art performance~\cite{jain_etal-2024}. DocRE-CLiP is the leading graph-based model but cannot be reproduced due to the absence of specific files in the shared GitHub repository. The other graph-based models are outperformed by the leading sequence-based model; hence, we do not consider them in our evaluation. Moreover, graph-based methods cannot effectively scale to the size of distantly supervised datasets.

\textbf{Sequence-based models} treat documents as a sequence of tokens and learn contextual representations for all entity pairs. Early approaches using CNNs and LSTMs struggled with long-range dependencies~\cite{yao_etal-2019}. Transformer-based models leveraging contextual embeddings, such as BERT, markedly improve the task and are now standard in DocRE~\cite{devlin-etal-2019-bert}.

A key transformer-based model is \textbf{ATLOP}, which introduces adaptive thresholding for entity-dependent multi-label classification, and localized context pooling for improved predictions~\cite{zhou_etal-2021}. KD-DocRE extends ATLOP by addressing class imbalance via adaptive focal loss and leveraging knowledge distillation in a teacher-student paradigm~\cite{tan_etal-2022a}. \citet{wang_etal-2024} frames \ac{DocRE} as a metric learning problem to learn similar representations for entity embeddings and ground truth relations. To enhance robustness, \citet{duan_etal-2025} proposes COMM, a two-stage model 
employing Concentrated Margin Maximization to adaptively adjust decision boundaries.
COMM is an enhancement strategy that can be applied to any off-the-shelf \ac{DocRE} models. Thus, this work is orthogonal to \ac{DOREMI} and can be integrated into the pipeline as future work.
\textbf{DREEAM} follows the same architecture as KD-DocRE and introduces an evidence-aware distillation mechanism that better guides learning using noisy \ac{DS} data, achieving state-of-the-art performance in the task~\cite{ma_etal-2023}. 
Other transformer-based or evidence-oriented models, such as SAIS~\cite{xiao_etal-2022} and EIDER~\cite{xie_etal-2022}, exhibit inferior results and suffer from scalability issues due to reliance on manual datasets. As DREEAM has shown superior performance over all these systems, we do not consider them as baselines for the experimental evaluation. 

\textbf{The long-tail problem in DocRE} has been addressed by three methods:
ERA~\cite{du_etal-2022}, PRISM~\cite{choiEtAl2023}, DocRE-Co-Occur~\cite{han_etal-2024}. ERA leverages a relation augmentation mechanism to enhance the entity pair representation by applying a random mask on pooled context representation. However, ERA shows inferior performance compared to DREEAM and long-tail results are inferior to DocRE-Co-Occur. PRiSM is a calibration-based approach that learns to adapt logits based on relation and entity pairs' embeddings similarity; it improves ~$0.5\%$ on ATLOP and is inferior to DREEAM. DocRE-Co-Occur introduces the concept of relation correlations, leveraging the co-occurrence patterns of relations to transfer knowledge from data-rich to data-scarce relations. By modeling these correlations through relation embeddings and incorporating auxiliary co-occurrence coarse and fine-grained predictions, DocRE-Co-Occur enhances the model's ability to extract rare relations. However, the performance of DocRE-Co-Occur is inferior to that of DREEAM, even considering only long-tail relations.
.

\textbf{Addressing noise in \ac{DS} datasets} is critical and could help balance long-tail relations via the injection of additional training data.
\citet{xiao_etal-2020} mitigate noise using multi-task learning across mention-entity matching, relation detection, and fact alignment. 
\citet{zhang_etal-2023} focuses on relational reasoning and proposes a self-distillation framework with a multihead attention unit that models common reasoning patterns.
\citet{li_etal-2023} proposes integrating a \ac{LLM} and a \ac{NLI} module to generate relation triples and perform data augmentation.
The current state-of-the-art and best-perfoming approach in label denoising is \textbf{UGDRE}, which applies Monte Carlo dropout to estimate pseudo-label uncertainty, selectively filtering low-confidence labels to improve training data quality~\cite{sun_etal-2023}. 

\begin{figure*}[t!]
    \centering
    \includegraphics[width=\linewidth]{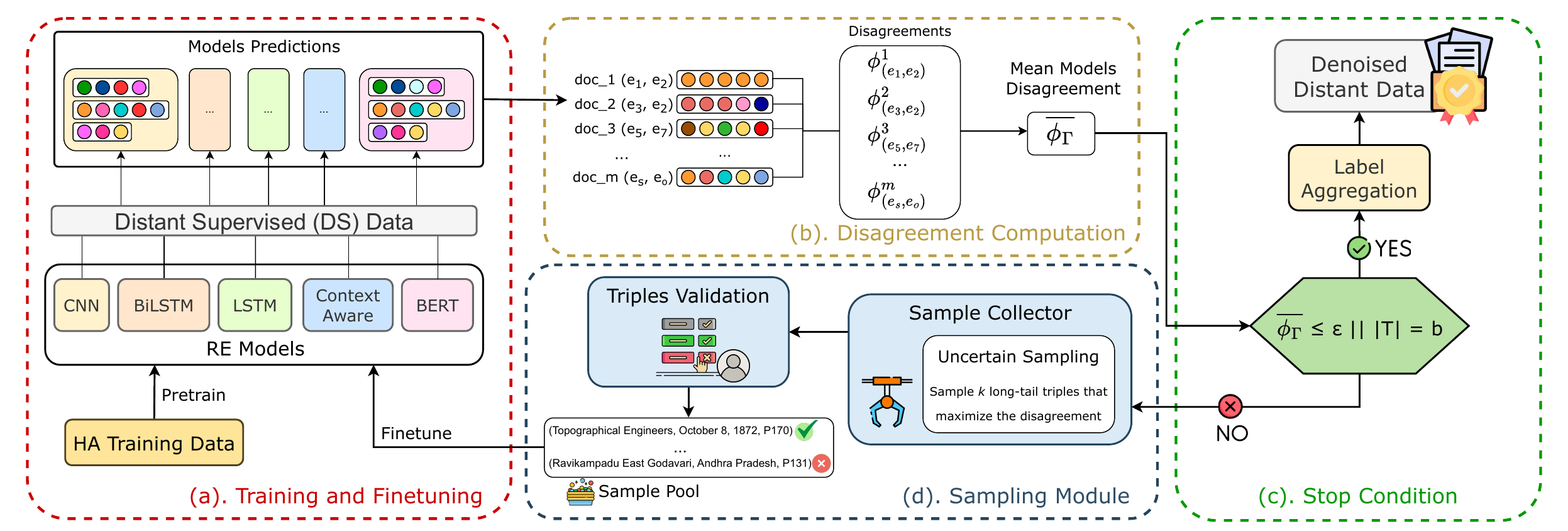}
    \caption{The DOREMI architecture including: (a) \emph{Training and Finetuning} of five models on long-tail relations; (b) \emph{Disagreement Computation} for each entity pair; (c) \emph{Stop Condition} checks if the disagreement is below a threshold or the annotation budget is exhausted; (d) \emph{Sampling Module} samples $k$ high-disagreement triples to be annotated.
}
    \label{fig:doremi}
\end{figure*}

In summary, we focus on \textbf{ATLOP}, \textbf{DREEAM}, and \textbf{UGDRE} as our primary baselines due to their strong performance, scalability, and representativeness of the best practices in modeling and data denoising. Other approaches are either incorporated within these models or perform worse and are therefore omitted from further comparison.

\section{DOREMI}
\label{sec:methodology}

\ac{DOREMI} aims to enhance the prediction of long-tail relations by selectively enriching the training data with minimal human-annotated triples. Given the high cost of expert annotations, we annotate as few examples as possible while maximizing their impact on improving long-tail performance. Figure~\ref{fig:doremi} shows the main computational blocks involved in \ac{DOREMI}. Table~\ref{tab:notation} reports the main symbols and notation used throughout the paper.

\begin{table}[th!]
    \centering
    \begin{tabularx}{\columnwidth}{l|X}
    \toprule
        $\Gamma$ & Set of core models \\
        $\gamma_i$ & i-th core model in $\Gamma$ \\
        $\phi_\Gamma(s,o)$ & Disagreement between models in $\Gamma$ for the entity pair $(s,o)$ \\
        $\overline{\phi}_{\Gamma}$ & Mean disagreement between models in $\Gamma$ for all entity pairs in the dataset \\
        $\epsilon$ & Disagreement threshold \\
        $k$ & Sample size \\
        $b$ & Annotation budget \\
        $\tau$ & Label aggregation threshold \\
    \bottomrule
    \end{tabularx}
    \caption{Notation table. We report the symbols used throughout the paper and in the figures to define the key elements of \ac{DOREMI}.}
    \label{tab:notation}
\end{table}

DOREMI maintains a pool \(\Gamma\) of \(n\) diverse \ac{DocRE} core models, each fine‐tuned on the available Human-Annotated ($HA$) training data (Figure~\ref{fig:doremi}a). These models then predict relations over the noisy \ac{DS} dataset and, for each entity pair $(s,o)$, these predicts are used to compute the disagreement  ($\phi_{\Gamma}(s,o)$) among models. The disagreement across all pairs is then averaged $\overline{\phi}_{\Gamma}$ and evaluated against a stopping criterion (Figure~\ref{fig:doremi}b).
Under the hypothesis that higher inter‐model disagreement indicates greater annotation utility, we stop annotating when \(\overline{\phi}_{\Gamma} \leq \varepsilon\) or when the annotation budget \(b\) is over. At that point, we aggregate model outputs into the final \ac{DDS} (Figure~\ref{fig:doremi}c). Otherwise, we sample the $k$ pairs with maximum disagreement for manual labeling, augment the training set, and repeat the cycle of model training and uncertainty‐driven sampling (Figure~\ref{fig:doremi}a,d). 

This section is organized as follows. Subsection~\ref{subsec:disagreement} details how disagreement is computed in \ac{DOREMI} to select entity pairs for annotation and to determine the stopping condition. Subsection~\ref{subsec:distant-constr} defines the label aggregation criteria employed by \ac{DOREMI} to build the \ac{DDS} in the final iteration. To conclude, Subsection~\ref{subsec:iter-training} describes the iterative training procedure followed by \ac{DOREMI}.
 
\subsection{Disagreement Computation}
\label{subsec:disagreement}
In the following, we define how we compute the disagreement between models in \ac{DOREMI}. The disagreement between models is employed to assess the stopping condition (Figure~\ref{fig:doremi}c) and select the entity pairs to annotate (Figure~\ref{fig:doremi}d).
\citet{le_etal-2024} demonstrated that sampling by model disagreement outperforms single‐model confidence sampling.  Inspired by this finding, we posit that inter‐model disagreement effectively identifies Hard‐To‐Classify examples in \ac{DocRE}.  Below, we formalize our disagreement measure using prediction probabilities produced by the models.

Let $\Gamma = \{\gamma_i\}^{n}_{i = 1}$ be a set of $n$ independently trained \ac{DocRE} models.  
For a candidate triple $(s, r, o)$, denote by $\mathds{1}_{\gamma_i}(r|(s,o))$ the indicator function returning $1$ if model $\gamma_i$ predicts relation $r$ between $s$ and $o$, and $0$ otherwise. 
Conversely, we write $\mathds{1}_{\gamma'_i}(r|(s,o))=1-\mathds{1}_{\gamma_i}(r|(s,o))$ for the event ``no relation $r$''.  When considering a single relation $r$, the models agree if they all predict $r$ or all predict ``no relation $r$''. Therefore, the probability of disagreement on $r$ can be computed as:

\begin{equation}
\label{eqn:initial}
\scalebox{0.9}{$
\begin{aligned}
\phi_{\Gamma}(r|(s,o)) = 1 - \Big[ 
    &\; P\left(\bigcap_{i=1}^n \mathds{1}_{\gamma_i}(r|(s,o))\right) \\
    +\; & P\left(\bigcap_{i=1}^n \mathds{1}_{\gamma'_i}(r|(s,o))\right) 
\Big]
\end{aligned}
$}
\end{equation}

Given that model predictions are independent, the probability that all models predict the relation can be expressed as the product of the individual model probabilities.

We denote $P(\mathds{1}_{\gamma_i}(r|(s,o))$ as $p_{\gamma_i}(r|(s,o)) = P\bigl(\gamma_i \text{ predicts relation }r\mid s,o\bigr)$. 
Thus, the probability that a model does not predict the relation is $1-p_{\gamma_i}(r|(s,o))$. Substituting into Equation~\ref{eqn:initial}, the disagreement is
\begin{equation}
\scalebox{0.9}{$
\begin{aligned}
\phi_{\Gamma}(r \mid (s,o)) = 1 - \Big[ 
    &\prod_{i=1}^n p_{\gamma_i}(r|(s,o)) \\
    +\; &\prod_{i=1}^n \left(1 - p_{\gamma_i}(r|(s,o))\right)
\Big]
\end{aligned}
$}
\end{equation}
In \ac{DocRE}, each of the $R$ target relations may hold independently for a given entity pair $(s,o)$, resulting in a multi‐label classification problem. Assuming independence across relations, the overall disagreement on the relation set $R$ for the entity pair $(s, o)$ is computed as the product of the individual per‐relation disagreements:
\begin{equation}
    \label{eqn:agreement}
    \phi_{\Gamma}(s,o) = \prod_{r \in R} \phi_{\Gamma}(r|(s,o)) 
\end{equation}

Since the disagreement $\phi_{\Gamma}{(r|(s,o))}$ lies in $[0,1]$, we apply a logarithmic transformation to amplify small differences and enhance interpretability. To ensure the logarithm is well-defined, we add a small, scale-invariant constant $\delta > 0$ to shift the values away from zero.\footnote{$\delta$ can be arbitrarily small; its value does not affect the approach. Hence, we omit $\delta$ from the equations.} This transformation maps disagreement values into the range $(-\infty, \delta]$, with complete disagreement corresponding to $\delta$:
\begin{equation}
    \psi_{\Gamma}(s,o) =  \sum_{r \in R} \log \phi_{\Gamma}(r|(s,o))
\end{equation}
Based on it, we define the sampling criterion for Hard‐To‐Classify instances as selecting the top‐$k$ entity pairs with the highest disagreement:
\begin{equation}
    \label{eqn:sc}
    \arg \max_{(s,o)} \psi_{\Gamma}(s, o)
\end{equation}
This strategy prioritizes entity pairs $(s, o)$ for which the model ensemble $\Gamma$ exhibits the highest uncertainty, thereby maximizing the expected benefit of manual annotation for each selected instance.

Since \ac{DOREMI} is tailored for target long-tail relations, the disagreement computation is limited to \emph{candidate long-tail triples} -- that is, $(s, o)$ pairs for which at least one core model predicts a long-tail relation. This focus ensures that annotation efforts are directed toward instances with the greatest potential to improve long-tail predictions. 

\subsection{Denoised Distant Dataset Construction}
\label{subsec:distant-constr}
Once the stopping condition is satisfied, we aggregate the core models predictions to construct the denoised distant dataset (figure~\ref{fig:doremi}c). This subsection describes the label aggregation criteria exploited by \ac{DOREMI}.
The overall mean disagreement $\overline{\phi}_\Gamma$ is computed by averaging $\phi_\Gamma(s,o)$ over all candidate pairs in the distantly supervised dataset.
Once $\overline{\phi}_{\Gamma} \le \varepsilon$ or the annotation budget $b$ is exhausted, the active learning loop ends. 
We then aggregate predictions from the best iteration of each model -- based on the highest long‐tail \textsc{F1} on the development set -- to construct the \ac{DDS}.

To maximize coverage while maintaining label quality in the \ac{DDS}, we adopt a selective retention strategy that filters out highly uncertain relations. Specifically, for each entity pair $(s, o)$ and relation $r$, we include $r$ in the final label set only if at least one model predicts it with high confidence:
\begin{equation*}
    \max_{i=1,\dots,n} p_{\gamma_i}(r | (s,o)) \;>\; \tau,
\end{equation*}
where $\tau$ is a confidence threshold. This approach acts as a precision-preserving filter, ensuring that only relations supported by strong evidence from at least one model are retained. As a result, we balance broad relational coverage with robustness, reducing noise without overly sacrificing recall. Thanks to this approach, we managed to retain around $60\%$ of the \ac{DDS} instances, while discarding the remaining.

\begin{algorithm}[t!]
\footnotesize
\caption{Iterative training for long-tail DocRE} 
\label{alg:doremi}
\begin{algorithmic}[1]
\small
\Require Human-annotated training data $HA$, distant training data $DS$, disagreement threshold $\varepsilon$, DocRE models $\Gamma = \gamma_1, .., \gamma_n$, sample size $k$, budget $b$
\Ensure Denoised distantly supervised data $DDS$

\For{\textbf{each} $\gamma_{i}$ \textbf{in} $\Gamma$}
\Comment{Pretrain the models (see Figure~\ref{fig:doremi}a)} 
    \State $\gamma^{pretrain}_i \gets \texttt{train}(\gamma_i, HA)$
    \State $\gamma^{prev}_i \gets \gamma^{pretrain}_i$
    \State $DDS^i \gets \texttt{predict}(\gamma_i^{pretrain}, DS)$
\EndFor
\Comment{Disagreement computation (see Figure~\ref{fig:doremi}b)} 
\State $\overline{\phi}_{\Gamma} \gets \texttt{mean}(\{\phi_{\Gamma}(s,o) : (s,o) \in DS \})$, $T \gets \emptyset$ 
\While{$\overline{\phi}_{\Gamma} > \varepsilon $ \textbf{and} $ |T| < b$}
    \Comment{Loop until stop condition is satisfied (see Figure~\ref{fig:doremi}c)}
    \State $S \gets$ Sample $k$ candidate long-tail triples with highest disagreement from $\{DDS^1, .., DDS^n\}$ \Comment{Sampling and annotation (see Figure~\ref{fig:doremi}d)}
    \State $SA \gets \texttt{annotate}(S)$
    \State $T \gets T \cup SA$
    \For{\textbf{each} $\gamma_{i}$ \textbf{in} $\Gamma$} \Comment{Finetune the models (see Figure~\ref{fig:doremi}a)}
        \small \State $\gamma^{finetune}_i \gets \texttt{finetune}(\gamma_i^{prev}, HA \cup T)$
        \State $\gamma^{prev}_i \gets \gamma^{finetune}_i$
        \State $DDS^i \gets \texttt{predict}(\gamma_i^{finetune}, DS)$
    \EndFor 
    \Comment{Disagreement computation (see Figure~\ref{fig:doremi}b)} 
    \State $\overline{\phi}_{\Gamma} \gets \texttt{mean}(\{{\phi_{\Gamma}(s,o)} : (s,o) \in DS \})$
\EndWhile \Comment{Stop condition reached (see Figure~\ref{fig:doremi}c)}
\For{\textbf{each} $\gamma_{i}$ \textbf{in} $\Gamma$} 
    \State $DSS^i \gets \texttt{predict}(\gamma_i^{finetune}, DS)$
\EndFor
\State $DDS \gets \texttt{merge}(DDS^1, .., DDS^n)$ \Comment{Label aggregation (see Figure~\ref{fig:doremi}c)}
\end{algorithmic}
\end{algorithm}

\begin{figure}[th!]
    \centering
    \includegraphics[width=\linewidth]{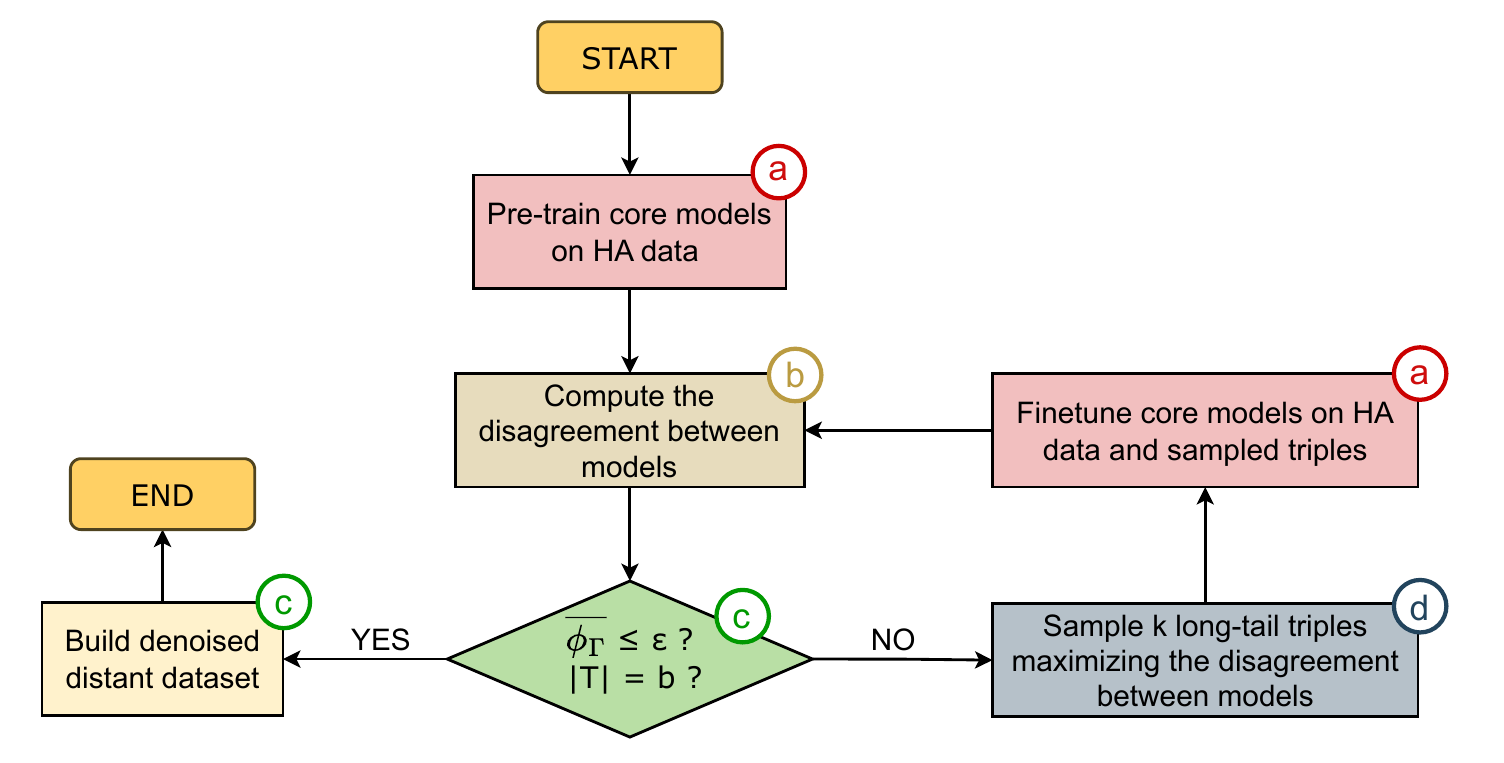}
    \caption{DOREMI flowchart. Each process or decision is linked to the corresponding computational blocks defined in Figure~\ref{fig:doremi}.}
    \label{fig:flowcart}
\end{figure}

\subsection{Iterative Training Procedure}
\label{subsec:iter-training}
This subsection details the iterative training procedure followed by \ac{DOREMI}. The pseudocode is reported in Algorithm~\ref{alg:doremi}. To enhance clarity, Figure~\ref{fig:flowcart} reports the flowchart of \ac{DOREMI}. In input, DOREMI takes the human-annotated ($HA$) and distantly supervised ($DS$) training data, a disagreement threshold ($\varepsilon$) for the stopping criterion, a set of \ac{DocRE} models $\Gamma = \{\gamma_i\}^{n}_{i = 1}$, a sample size ($k$), and an annotation budget ($b$).  

In lines~1-5,  each \ac{DocRE} model $\gamma_i$ is pretrained on the $HA$ training set, stored, and used to label the $DS$ data. The predictions are then used to compute the model disagreement for each entity pair $(s, o)$, which in turn is used to calculate the preliminary mean disagreement $\overline{\phi}_{\Gamma}$ (line 6), averaging over all the pair disagreements between the models.  

Iterations (lines 7-17) continue as long as the average disagreement among the models is above threshold ($\overline{\phi}_{\Gamma} > \varepsilon$) and annotation budget is available ($|T| < b$). Specifically, DOREMI samples $k$ candidate long-tail triples based on the highest disagreement (line 8), which are then annotated and stored in the sample pool $T$ (lines 9,10). Each model $\gamma_i$ is fine-tuned on the combination of $HA$ input data and newly annotated data $T$ (line 12). The fine-tuned model is stored to serve as a starting point for the next iteration (line 13). Predictions are made on the $DS$ data using the fine-tuned model, resulting in the set $DSS^i$ (line 14). Once predictions are obtained from all models, they are used to compute the model disagreement for each entity pair, as well as the \emph{current iteration} average disagreement (line 16). 

After the loop concludes, each model $\gamma_i$ makes a final round of predictions on $DS$ data (line 19). These predictions are aggregated across models to generate the \ac{DDS} (line 21).

\section{Experimental Setup}
\label{sec:setup}
This section describes the experimental setup exploited to evaluate \ac{DOREMI}. In particular, Subsection~\ref{subsec:core-mods} reports the core models used to predict the relations in the noisy \ac{DS} data and finetuned at each iteration. Subsection~\ref{subsec:data} reports the dataset used for training and testing \ac{DOREMI}. To conclude, Subsection~\ref{subsec:baselines} describes the baselines chosen to evaluate the effectiveness of \ac{DOREMI} denoising.

The implementation of \ac{DOREMI} is available in GitHub at: \url{https://github.com/mntlra/DOREMI}. Together with the source code, we release some Python scripts to reproduce the \ac{DOREMI} iterative training procedure. Table~\ref{tab:params} summarizes the hyperparameters used in \ac{DOREMI}.

\begin{table}[th!]
    \centering
    \begin{tabular}{l|c}
    \toprule
    \multicolumn{2}{l}{\textbf{(a) Training with DocRED}} \\[0.5em]
       Sample size $k$  & 100 \\
       Annotation budget $b$  & 400 \\
    \midrule
    \multicolumn{2}{l}{\textbf{(b) Training with Re-DocRED}} \\[0.5em]
        Sample size $k$  & 300 \\
       Annotation budget $b$  & 1,200 \\
    \midrule
    \multicolumn{2}{l}{\textbf{(c) Common parameters}} \\[0.5em]
    Label aggregation threshold $\tau$ & 0.7 \\
    \bottomrule
    \end{tabular}
    \caption{DOREMI experimental setting. The table summarizes the hyperparameters exploited in DOREMI.}
    \label{tab:params}
\end{table}

\subsection{Core Models}
\label{subsec:core-mods}

We chose five core models ($n=5$) for DOREMI due to two primary reasons. Firstly, using an odd number of models avoids tie cases during aggregation. Secondly, we adopted the core models originally used in DocRED~\cite{yao_etal-2019}, the first benchmark dataset for \ac{DocRE}. Specifically, DocRED employs CNN, LSTM, BiLSTM, and ContextAware (based on word2vec). the code to train the core models exploited in DocRED is publicly accessible in GitHub.~\footnote{\url{https://github.com/thunlp/DocRED/}} Given the widespread adoption of contextual embeddings, in particular BERT~\cite{devlin-etal-2019-bert}, and their strong capability to model long-range contextual dependencies for relation prediction, we incorporated it into \ac{DOREMI} as the fifth core model. We developed our own core model exploiting BERT. the implementation is based on DREEAM~\footnote{\url{https://github.com/YoumiMa/dreeam}} and ATLOP~\footnote{\url{https://github.com/wzhouad/ATLOP}} GitHub repositories.
A higher number of models serves a proxy to enhance diversity between the models, as in the case of $n=5$. We decided to not include more recent models tailored for \ac{DocRE} because most of them rely on either BERT or RoBERTa as the backbone transformer architecture~\cite{han_etal-2024,choiEtAl2023,zhou_etal-2021,ma_etal-2023,tan_etal-2022a,duan_etal-2025,xiao_etal-2022,xie_etal-2022}. Thus, we do not expect these models to meaningfully increase diversity, as they are likely to exhibit score distributions similar to BERT (reported in Figure~\ref{fig:score-distribution}).

During training, we followed the same hyperparameters adopted in~\cite{yao_etal-2019}. Table~\ref{tab:core-params} reports the experimental setting used in \ac{DOREMI}. All baselines were optimized using Adam. For the BERT core mode, we adopted the same hyperparameter settings of DREEAM~\cite{ma_etal-2023}.

\begin{table}[th!]
    \centering
    \begin{tabular}{l|c}
    \toprule
    \multicolumn{2}{l}{\textbf{(a) DocRED core models}} \\[0.5em]
      Batch size   &  40 \\
      Pre-training epochs & 200 \\
      Finetuning epochs & 70 \\
      Learning rate & 0.001 \\
      CNN hidden size & 200 \\
      CNN window size & 3 \\
      CNN dropout rate & 0.5 \\
      LSTM hidden size & 128 \\
      LSTM dropout rate & 0.2 \\
      Word embedding dimension & 100 \\
      Entity type embedding dimension & 20 \\
      Coreference embedding dimension & 20 \\
      Distance embedding dimension & 20 \\
    \midrule
    \multicolumn{2}{l}{\textbf{(b) BERT core model}} \\[0.5em]
      Batch size & 4 \\
      Pre-training epochs & 30 \\
      Finetuning epochs & 10 \\
      Transformer learning rate & 3e-5 \\
      Classifier learning rate & 1e-4 \\
      Warmup ratio & 0.06 \\
      Maximum norm of the gradients & 1.0 \\
    \bottomrule
    \end{tabular}
    \caption{Core models hyperparameter settings. We employed the same training setting as DocRED~\cite{yao_etal-2019} for CNN, BiLSTM, LSTM, and ContextAware. the hyperparameters for BERT are taken from DREEAM implementation~\cite{ma_etal-2023}.}
    \label{tab:core-params}
\end{table}

\subsection{Datasets} 
\label{subsec:data}
We used DOREMI with DocRED~\cite{yao_etal-2019} and Re-DocRED~\cite{tan_etal-2022b}, 
 a polished version of the former.
Despite known annotation issues in DocRED, it remains a widely used benchmark~\cite{zhang_etal-2025a,zhang_etal-2024,fan_etal-2024,qi_etal-2024,zhang_etal-2025b}. 
In the DocRED configuration, the core models are iteratively trained starting from the DocRED annotated training set, with the optimal iteration selected based on the highest long‐tail \textsc{F1} evaluated on the DocRED development set. The test dataset of DocRED is not publicly accessible; the dataset can be evaluated by submitting the models predictions to CodaLab~\footnote{\url{https://codalab.lisn.upsaclay.fr/competitions/365}}, which solely reports the relation extraction F1 and evidence F1 on the test set. Thus, we cannot perform a long-tail evaluation exploiting it. Given this dataset limitation, previous works on DocRE report the final performance on both the DocRED development and test datasets~\cite{han_etal-2024,choiEtAl2023,zhou_etal-2021,ma_etal-2023,tan_etal-2022a,duan_etal-2025,xiao_etal-2022,xie_etal-2022, zhang_etal-2024,zhang_etal-2025b}. Thus, we decided to perform a detailed evaluation on the DocRED development dataset and not change the dataset splitting to ensure compatibility with previous works. In this way, DOREMI results can be compared to any existing DocRE model exploiting DocRED. We also evaluated our approach on the DocRED test dataset and included the micro-averaged F1 score on the full dataset.

We define long-tail relations as those with less than $100$ instances in the training set (cf. Figure~\ref{fig:lbl-distribution}a), and set the sampling size $k=100$. To keep annotation costs minimal, we annotated only the $0.001\%$ ($b=400$) of the distantly supervised dataset. Therefore, at each iteration, the fine-tuning dataset grows by $0.3\%$. 
For Re-DocRED, we define long-tail relations as those having less than $300$ training instances (cf. Figure~\ref{fig:lbl-distribution}b). Since Re-DocRED contains roughly three times more triples than DocRED -- $85,932$ vs $38,180$ -- we scale both $k$ and $b$ proportionally, setting $k=300$ and $b=1,200$ ($0.003\%$). This proportional scaling ensures that the fine-tuning dataset grows by $0.3\%$ as for DocRED.

\begin{figure*}[t]
    \centering
    \includegraphics[width=\textwidth]{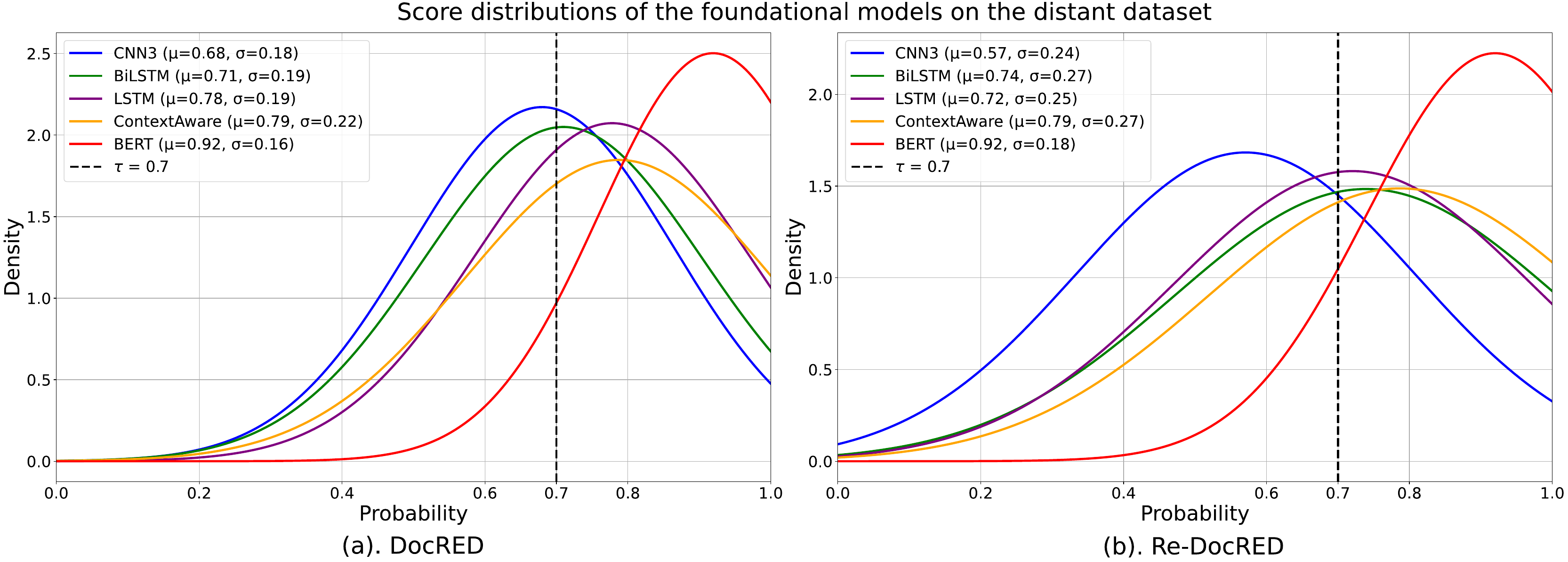}
    \caption{Score distributions of the core models trained using DocRED (a) and Re-DocRED (b) in the DocRED distant dataset.} 
    \label{fig:score-distribution}
\end{figure*}
Figure~\ref{fig:score-distribution} reports the score distribution of the distant triple predicted by the core models trained using DocRED (Figure~\ref{fig:score-distribution}a) and Re-DocRED (Figure~\ref{fig:score-distribution}b). The distribution is skewed towards high confidence for both datasets, with most models exhibiting a mean confidence score ($\mu$) around $0.7$. Based on this insight, we set the threshold $\tau = 0.7$ to retain only relations strongly supported by at least one model. This choice of $\tau$ ensures a robust balance between coverage and noise reduction, effectively acting as a precision-preserving filter without overly sacrificing recall.

Using this threshold during the aggregation step, \ac{DOREMI} retains $65.5\%$ of predicted relations on DocRED, producing a dataset with $1,704,471$ triples. On Re-DocRED, it preserves $61.5\%$ of relations, yielding $4,158,468$ triples.

\subsection{Baselines}
\label{subsec:baselines}
To assess the effectiveness of our \ac{DDS}, we train state-of-the-art \ac{DocRE} models from scratch on three datasets: (i) the original DocRED \ac{DS} dataset~\cite{yao_etal-2019}, (ii) the denoised UGDRE dataset~\cite{sun_etal-2023}, and (iii) the DOREMI \ac{DDS} (ours). DOREMI aims to improve performance in long-tail relations; hence, it can complement approaches focusing on frequent relations, such as UGDRE. To demonstrate this, we evaluate each model on a hybrid dataset, denoted D+U, which combines DOREMI long-tail predictions with UGDRE frequent relations annotations.

We adopt two transformer-based architectures as baselines: ATLOP~\cite{zhou_etal-2021} and DREEAM~\cite{ma_etal-2023}. Both approaches support multiple transformer backbones; we report results using BERT-base and RoBERTa-large. DREEAM follows a teacher–student paradigm, where the student model is first trained on \ac{DS} data (called ``\emph{self-training}") and then fine-tuned on human-annotated data. Since our focus is on evaluating the quality of distantly supervised datasets, we report the performance of the DREEAM student model right after self-training. 
These models have been selected due to their strong empirical performance and the availability of open-source implementations (cf. Section~\ref{sec:related}). The implementation of ATLOP~\footnote{\url{https://github.com/wzhouad/ATLOP}} and DREEAM~\footnote{\url{https://github.com/YoumiMa/dreeam}} is publicly available in GitHub. Thus, we train the two models from scratch, leveraging the experimental setup and hyperparameter settings of the corresponding reference papers. We share the scripts to train DREEAM and ATLOP using different distant datasets in the \texttt{repro} directory of our GitHub repository.~\footnote{\url{https://github.com/mntlra/DOREMI/tree/main/repro}}
Although UGDRE~\cite{sun_etal-2023} also provides a competitive \ac{DocRE} model along with the denoised dataset, we could not reproduce their results due to deprecated dependencies. 
Model performance is reported using micro-averaged metrics, as this is the standard evaluation approach commonly adopted in \ac{DocRE} literature.

\section{Experimental Results}
\label{sec:results}
This section reports \ac{DOREMI} results compared to a state-of-the-art approach for denoising \ac{DS} data. In particular, subsection~\ref{subsec:docred} reports the performance of ATLOP and DREEAM evaluated on the DocRED development dataset. In this configuretion, \ac{DOREMI} is trained and finetuned on the DocRED training dataset. Subsection~\ref{subsec:redocred} reports the performance of ATLOP and DREEAM evaluated on the Re-DocRED test dataset. In this configuretion, \ac{DOREMI} is trained and finetuned on the Re-DocRED training dataset. Since Re-DocRED is three times larger than DocRED, we consider long-tail relations those having less than $300$ training examples. Relations having less than $100$ training examples represent the "\emph{extreme long-tail}". Extreme long-tail performances are reported in Subsection~\ref{subsec:long-tail}.

\subsection{DocRED}
\label{subsec:docred}

Table~\ref{tab:results-docred} reports the performance of ATLOP and DREEAM trained on different distant datasets and evaluated on the DocRED development (dev) dataset. We rely on the dev dataset because the test set is not publicly available and thus prevents long-tail analysis.

\begin{table*}[t!]
\caption{Performance of \ac{DocRE} models trained on distant datasets: DocRED, UGDRE, DOREMI \textbf{(ours)}, and D+U (DOREMI for long-tail, UGDRE for frequent relations). Results are on the \textbf{DocRED dev set}, except for column ``Test-F1", which reports the micro-F1 for the full dataset on the \textbf{DocRED test set}. Long-tail refers to relations with $<100$ training instances. Best and second-best results are bolded and underlined.}

\label{tab:results-docred}
\resizebox{\textwidth}{!}{%
\begin{tabular}{ll|c|ccccc|ccccc}
\toprule
\multirow{2}{*}{Model} & \multirow{2}{*}{Distant Data} & \multicolumn{6}{c|}{Full Dataset} & \multicolumn{5}{c}{Long-Tail Triples} \\ 
\multicolumn{1}{l}{} & & Test-F1 & Precision & Ign Prec &  Recall & F1 & Ign F1 & Precision & Ign Prec & Recall & F1 & Ign F1 \\ 
\midrule
\multirow{4}{*}{\begin{tabular}[c]{@{}l@{}}ATLOP \\ BERT\end{tabular}} & DocRED & 0.5326 & 0.5139 & 0.3346 & 0.5725 & 0.5416 & 0.4223 & 0.1917 & 0.0857 & \underline{0.3839} & 0.2557 & 0.1401 \\
 & UGDRE & 0.5797 & 0.5362 & 0.3711 & \textbf{0.6725} & \underline{0.5967} & 0.4783 & 0.2332 & 0.1293 &  \textbf{0.4013} & \textbf{0.2950} & 0.1956 \\
 & \cellcolor{ours} DOREMI & \cellcolor{ours} \underline{0.5883} & \cellcolor{ours} \textbf{0.5722} & \cellcolor{ours} \textbf{0.4384} & \cellcolor{ours}  0.6160 & \cellcolor{ours} 0.5933 & \cellcolor{ours} \textbf{0.5123} & \cellcolor{ours} \underline{0.3634} & \cellcolor{ours} \underline{0.2701} & \cellcolor{ours} 0.1946 & \cellcolor{ours} 0.2535 & \cellcolor{ours} \underline{0.2262} \\
 & \cellcolor{ours} D+U & \cellcolor{ours} \textbf{0.5901} & \cellcolor{ours} \underline{0.5600} & \cellcolor{ours} \underline{0.3930} & \cellcolor{ours} \underline{0.6416} & \cellcolor{ours} \textbf{0.5980} & \cellcolor{ours} \underline{0.4874} & \cellcolor{ours} \textbf{0.3807} & \cellcolor{ours} \textbf{0.2744} & \cellcolor{ours} 0.2336 & \cellcolor{ours} \underline{0.2895} & \cellcolor{ours} \textbf{0.2523} \\ 
\midrule
\multirow{4}{*}{\begin{tabular}[c]{@{}l@{}}ATLOP \\ RoBERTa\end{tabular}} & DocRED & 0.5409 & 0.5094 & 0.3345 & 0.5912 & 0.5472 & 0.4272 & 0.1913 & 0.0876 & \underline{0.4067} & 0.2602 & 0.1442 \\
 & UGDRE & 0.5840 & 0.5350 & 0.3746 & \textbf{0.6810} & 0.5992 & 0.4834 & 0.2387 & 0.1399 & \textbf{0.4604} & \textbf{0.3144} & 0.2146 \\
 & \cellcolor{ours} DOREMI & \cellcolor{ours} \underline{0.5909} & \cellcolor{ours} \textbf{0.5788} & \cellcolor{ours} \textbf{0.4453} & \cellcolor{ours} 0.6242 & \cellcolor{ours} \underline{0.6006} & \cellcolor{ours} \textbf{0.5198} & \cellcolor{ours} \textbf{0.4200} & \cellcolor{ours} \textbf{0.3324} & \cellcolor{ours} 0.2362 & \cellcolor{ours} \underline{0.3024} & \cellcolor{ours} \textbf{0.2762} \\
 & \cellcolor{ours} D+U & \cellcolor{ours} \textbf{0.5998} & \cellcolor{ours} \underline{0.5657} & \cellcolor{ours} \underline{0.4063} & \cellcolor{ours} \underline{0.6528} & \cellcolor{ours} \textbf{0.6061} & \cellcolor{ours} \underline{0.5009} & \cellcolor{ours} \underline{0.3713} & \cellcolor{ours} \underline{0.2737} & \cellcolor{ours} 0.2188 & \cellcolor{ours} 0.2753 & \cellcolor{ours} \underline{0.2432} \\ 
\specialrule{1pt}{0pt}{2pt}
\multirow{4}{*}{\begin{tabular}[c]{@{}l@{}}DREEAM \\ BERT\end{tabular}} & DocRED & 0.5306 & 0.5602 & 0.3905 & 0.5036 & 0.5304 & 0.4399 & 0.1714 & 0.0862 & 0.2443 & 0.2013 & 0.1274 \\
 & UGDRE & 0.5966 & \underline{0.5915} & 0.4366 & \underline{0.6138} & 0.6025 & 0.5102 & 0.2649 & 0.1652 & \textbf{0.3154} & 0.2880 & 0.2168 \\
 & \cellcolor{ours} DOREMI & \cellcolor{ours} \textbf{0.6095} & \cellcolor{ours} 0.5765 & \cellcolor{ours} \underline{0.4433} & \cellcolor{ours} \textbf{0.6341} & \cellcolor{ours} \underline{0.6039} & \cellcolor{ours} \textbf{0.5218} & \cellcolor{ours} \underline{0.3928} & \cellcolor{ours} \underline{0.3058} & \cellcolor{ours} 0.2483 & \cellcolor{ours} \underline{0.3043} & \cellcolor{ours} \underline{0.2741} \\
 & \cellcolor{ours} D+U & \cellcolor{ours} \underline{0.6011} & \cellcolor{ours} \textbf{0.6108} & \cellcolor{ours} \textbf{0.4578} & \cellcolor{ours} 0.5991 & \cellcolor{ours} \textbf{0.6049} & \cellcolor{ours} \underline{0.5190} & \cellcolor{ours} \textbf{0.4387} & \cellcolor{ours} \textbf{0.3444} & \cellcolor{ours} \underline{0.2497} & \cellcolor{ours} \textbf{0.3182} & \cellcolor{ours} \textbf{0.2895} \\ 
\midrule
\multirow{4}{*}{\begin{tabular}[c]{@{}l@{}}DREEAM \\ RoBERTa\end{tabular}}& DocRED & 0.5459 & 0.5582 & 0.3890 & 0.5370 & 0.5474 & 0.4512 & 0.1834 & 0.0965 & \underline{0.3047} & 0.2289 & 0.1466\\
 & UGDRE & 0.6064 & 0.5853 & \underline{0.4335} & \underline{0.6481} & 0.6151 & 0.5195 & 0.2656 & 0.1723 & \textbf{0.3544} & 0.3036 & 0.2319 \\
  & \cellcolor{ours} DOREMI & \cellcolor{ours} \textbf{0.6198} & \cellcolor{ours} \underline{0.5904} & \cellcolor{ours} \textbf{0.4564} & \cellcolor{ours} \textbf{0.6513} & \cellcolor{ours} \textbf{0.6194} & \cellcolor{ours} \textbf{0.5367} & \cellcolor{ours} \textbf{0.4332} & \cellcolor{ours} \textbf{0.3492} & \cellcolor{ours} 0.2523 & \cellcolor{ours} \textbf{0.3189} & \cellcolor{ours} \textbf{0.2930} \\
 & \cellcolor{ours} D+U & \cellcolor{ours} \underline{0.6144} & \cellcolor{ours} \textbf{0.6090} & \cellcolor{ours} \textbf{0.4564} & \cellcolor{ours} 0.6287 & \cellcolor{ours} \underline{0.6187} & \cellcolor{ours} \underline{0.5289} & \cellcolor{ours} \underline{0.4226} & \cellcolor{ours} \underline{0.3261} & \cellcolor{ours} 0.2456 & \cellcolor{ours} \underline{0.3107} & \cellcolor{ours} \underline{0.2802} \\
\bottomrule
\end{tabular}%
}
\end{table*}

When trained on the \ac{DDS} generated by \ac{DOREMI}, all \ac{DocRE} models exhibit superior performance -- both overall and on long-tail relations. Compared to UGDRE, the current state-of-the-art in distant label denoising, \ac{DOREMI} improves overall \textsc{precision} by up to $+8.2\%$ and \textsc{F1} by up to $+0.7\%$. The improvement of \ac{DOREMI} over UGDRE in terms of \textsc{F1} score is statistically significant for three out of four models, as determined by a paired t-test ($p < 0.01$). The overall improvement of \ac{DOREMI} over UGDRE in terms of \textsc{F1} score is confirmed by the performance on the DocRED test dataset (column ``Test-F1" of Table~\ref{tab:results-docred}).
In long-tail prediction, \ac{DOREMI} achieves a \textsc{precision} of $43.3\%$, representing a relative gain of $+76.0\%$ over UGDRE with ATLOP-RoBERTa, and $+63.1\%$  with DREEAM-RoBERTa. 
We further evaluate performance using \textsc{ignPrecision} and \textsc{ignF1}, which exclude entity pairs observed during training. On these metrics, DocRED and UGDRE exhibit steep precision drops (nearly $-50\%$) on unseen long-tail triples. In contrast, \ac{DOREMI} delivers two to three times higher \textsc{ignPrecision} than UGDRE, scoring a performance gain of up to $+137.6\%$. 

Remarkably, these improvements are attained with 400 manual annotations -- just \textbf{0.001\%} of all candidate pairs -- underscoring the efficiency of DOREMI in constructing a high-quality corpus with minimal human effort.

Finally, in a hybrid configuration (D+U) that combines DOREMI labels for rare relations with UGDRE labels for frequent ones, D+U matches or exceeds UGDRE in most metrics in both the DocRED dev and test dataset. These findings indicate that iterative, disagreement-driven labeling yields stronger denoising capabilities than existing approaches and can be seamlessly integrated with existing frequency-based techniques.

\begin{table*}[t!]
\caption{Performance of \ac{DocRE} models trained on distantly supervised datasets: DocRED, UGDRE, DOREMI \textbf{(ours)}, and D+U (DOREMI for long-tail, UGDRE for frequent relations). Results are on the \textbf{Re-DocRED test set}; long-tail refers to relations with $<300$ training instances. Best and second-best results are bolded and underlined.} 

\label{tab:results-redocred}
\resizebox{\textwidth}{!}{%
\begin{tabular}{ll|ccccc|ccccc}
\toprule
\multirow{2}{*}{Model} & \multirow{2}{*}{Distant Data} & \multicolumn{5}{c|}{Full Dataset} & \multicolumn{5}{c}{Long-Tail Triples} \\ 
\multicolumn{1}{l}{} & & Precision & Ign Prec & Recall & F1 & Ign F1 & Precision & Ign Prec & Recall & F1 & Ign F1 \\ 
\midrule
\multirow{4}{*}{\begin{tabular}[c]{@{}l@{}}ATLOP \\ BERT\end{tabular}} & DocRED & 0.7353 & 0.5793 & 0.3000 & 0.4261 & 0.3953 & 0.3932 & 0.2610 & 0.2710 & 0.3209 & 0.2659 \\
 & UGDRE & \underline{0.8044} & \underline{0.7445} & \textbf{0.7007} & \textbf{0.7490} & \textbf{0.7220} & 0.6034 & \underline{0.5280} & \textbf{0.5513} & \textbf{0.5762} & \textbf{0.5394} \\ 
 & \cellcolor{ours} DOREMI & \cellcolor{ours} 0.7480 & \cellcolor{ours} 0.6297 & \cellcolor{ours} \underline{0.6914} & \cellcolor{ours} 0.7186 & \cellcolor{ours} 0.6591 & \cellcolor{ours} \textbf{0.6741} & \cellcolor{ours} \textbf{0.6061} & \cellcolor{ours} 0.4090 & \cellcolor{ours} 0.5091 & \cellcolor{ours} 0.4884 \\
 & \cellcolor{ours} D+U & \cellcolor{ours} \textbf{0.8340} & \cellcolor{ours} \textbf{0.7779} & \cellcolor{ours} 0.6580 & \cellcolor{ours} \underline{0.7356} & \cellcolor{ours} \underline{0.7129} & \cellcolor{ours} \underline{0.6087} & \cellcolor{ours} 0.5260 & \cellcolor{ours} \underline{0.4699} & \cellcolor{ours} \underline{0.5303} & \cellcolor{ours} \underline{0.4963} \\
\midrule
\multirow{4}{*}{\begin{tabular}[c]{@{}l@{}}ATLOP \\ RoBERTa\end{tabular}} & DocRED & 0.7412 & 0.6013 & 0.3241 & 0.4510 & 0.4212 & 0.4029 & 0.2885 & 0.3064 & 0.3480 & 0.2972 \\
 & UGDRE & \underline{0.8090} & \underline{0.7502} & \textbf{0.7087} & \textbf{0.7555} & \textbf{0.7288} & 0.5994 & 0.5248 & \textbf{0.5600} & \textbf{0.5791} & \textbf{0.5419} \\ 
 & \cellcolor{ours} DOREMI & \cellcolor{ours} 0.7486 & \cellcolor{ours} 0.6311 & \cellcolor{ours} \underline{0.7023} & \cellcolor{ours} 0.7247 & \cellcolor{ours} 0.6648 & \cellcolor{ours} \textbf{0.6715} & \cellcolor{ours} \textbf{0.6112} & \cellcolor{ours} 0.4020 & \cellcolor{ours} 0.5029 & \cellcolor{ours} 0.4850 \\
 & \cellcolor{ours} D+U & \cellcolor{ours} \textbf{0.8264} & \cellcolor{ours} \textbf{0.7707} & \cellcolor{ours} 0.6888 & \cellcolor{ours} \underline{0.7513} & \cellcolor{ours} \underline{0.7275} & \cellcolor{ours} \underline{0.6049} & \cellcolor{ours} \underline{0.5326} & \cellcolor{ours} \underline{0.5182} & \cellcolor{ours} \underline{0.5582} & \cellcolor{ours} \underline{0.5253} \\
\specialrule{1pt}{0pt}{2pt}
\multirow{4}{*}{\begin{tabular}[c]{@{}l@{}}DREEAM \\ BERT\end{tabular}} & DocRED & 0.8087 & 0.6791 & 0.2609 & 0.3945 & 0.3770 & 0.4429 & 0.3166 & 0.1939 & 0.2697 & 0.2405 \\
 & UGDRE & \underline{0.8265} & \underline{0.7740} & \underline{0.6966} & \textbf{0.7560} & \textbf{0.7333} & 0.6364 & 0.5729 & \textbf{0.5182} & \textbf{0.5713} & \textbf{0.5442} \\ 
 & \cellcolor{ours} DOREMI & \cellcolor{ours} 0.7701 & \cellcolor{ours} 0.6576 & \cellcolor{ours} \textbf{0.7037} & \cellcolor{ours} \underline{0.7354} & \cellcolor{ours} 0.6799 & \cellcolor{ours} \textbf{0.7250} & \cellcolor{ours} \textbf{0.6640} & \cellcolor{ours} 0.4210 & \cellcolor{ours} \underline{0.5326} & \cellcolor{ours} \underline{0.5153} \\
 & \cellcolor{ours} D+U & \cellcolor{ours} \textbf{0.8562} & \cellcolor{ours} \textbf{0.8086} & \cellcolor{ours} 0.6433 & \cellcolor{ours} 0.7347 & \cellcolor{ours} \underline{0.7166} & \cellcolor{ours} \underline{0.6642} & \cellcolor{ours} \underline{0.5992} & \cellcolor{ours} \underline{0.4340} & \cellcolor{ours} 0.5250 & \cellcolor{ours} 0.5034 \\ 
\midrule
\multirow{4}{*}{\begin{tabular}[c]{@{}l@{}}DREEAM \\ RoBERTa\end{tabular}} & DocRED & 0.8170 & 0.6964 & 0.2861 & 0.4238 & 0.4056 & 0.4672 & 0.3444 & 0.2357 & 0.3134 & 0.2799 \\
 & UGDRE & \underline{0.8464} & \underline{0.7996} & \underline{0.7252} & \textbf{0.7811} & \textbf{0.7606} & 0.6517 & 0.5901 & \textbf{0.5530} & \textbf{0.5983} & \textbf{0.5709} \\ 
 & \cellcolor{ours} DOREMI & \cellcolor{ours} 0.7939 & \cellcolor{ours} 0.6894 & \cellcolor{ours} \textbf{0.7267} & \cellcolor{ours} 0.7588 & \cellcolor{ours} 0.7075 & \cellcolor{ours} \textbf{0.7575} & \cellcolor{ours} \textbf{0.7032} & \cellcolor{ours} 0.4498 & \cellcolor{ours} 0.5644 & \cellcolor{ours} \underline{0.5486} \\
 & \cellcolor{ours} D+U & \cellcolor{ours} \textbf{0.8766} & \cellcolor{ours} \textbf{0.8354} & \cellcolor{ours} 0.6738 & \cellcolor{ours} \underline{0.7619} & \cellcolor{ours} \underline{0.7459} & \cellcolor{ours} \underline{0.6936} & \cellcolor{ours} \underline{0.6349} & \cellcolor{ours} \underline{0.4807} & \cellcolor{ours} \underline{0.5679} & \cellcolor{ours} 0.5472 \\ 
\bottomrule
\end{tabular}%
}
\end{table*}
\subsection{Re-DocRED}
\label{subsec:redocred}

Table~\ref{tab:results-redocred} presents the performance of ATLOP and DREEAM trained on different distant datasets and evaluated on the Re-DocRED test dataset. 
Models trained with the \ac{DOREMI} \ac{DDS} consistently outperform those employing the original DocRED \ac{DS} dataset and the UGDRE one in terms of long-tail \textsc{precision} and \emph{ignored} precision (\textsc{ignPrecision}) -- regardless of model architecture (ATLOP or DREEAM) and transformer backbone (BERT or RoBERTa). Compared to the UGDRE denoised dataset, training with the \ac{DOREMI} \ac{DDS} reaches a \textsc{precision} of $75.75\%$, representing a gain of $10.58$ points over UGDRE ($65.17\%$) with DREEAM-RoBERTa. Pronounced improvements are observed in the \emph{ignored} metrics, which exclude entity pairs observed during training. Compared to UGDRE, DOREMI achieves a $+19.2\%$ gain in long-tail \textsc{ignPrecision}, highlighting its superior capability of generalizing to new unseen entity pairs.

Considering the full dataset, \ac{DOREMI} achieves the highest recall on DREEAM for both transformer backbones, registering a statistically significant improvement of up to $+1.0\%$ compared to UGDRE ($p < 0.01$).
In the hybrid D+U setting, top performance is obtained for full precision-related metrics ($+3.7\%$ on UGDRE), indicating that DOREMI disagreement-driven supervision effectively complements existing denoising approaches.

These gains are achieved by annotating just \textbf{0.003\%} of the distant dataset, 
underscoring the efficiency of DOREMI in producing high-quality labels at minimal annotation cost. The number of triples annotated to achieve these results is $1,200$. This amounts to roughly $20$ annotator-hours of work, based on the annotation rate of one minute per triple that we observed during the annotation phase. This highlights its potential to significantly improve \ac{DocRE} and \ac{KBC}. 
The results for \emph{extreme long-tail triples}, defined as relations with less than $100$ examples in the Re-DocRED training dataset, reported in the next section, confirm the same trend.

\subsubsection{Extreme Long-Tail Relations}
\label{subsec:long-tail}
Re-DocRED long-tail triples have less than $300$ examples in the annotated training dataset. This section investigates the impact of the \ac{DOREMI} \ac{DDS} in the \emph{extreme long-tail triples}, i.e., triples with less than $100$ examples in the Re-DocRED annotated training dataset.
As shown in Table~\ref{tab:redocred-100}, \ac{DOREMI} shows superior results compared to the DocRED \ac{DS} dataset and UGDRE, especially in \textsc{precision} ($+83.2\%$ compared to UGDRE). When combined with UGDRE, i.e. dataset D+U, the dataset outperforms \ac{DOREMI} when considering DREEAM with BERT-base as a backbone and reports superior performance compared to the DocRED \ac{DS} dataset and UGDRE. The findings validate that \ac{DOREMI} can be effectively integrated with various systems to enhance performance further. 

\begin{table}[t!]
\caption{Micro-averaged performance on extreme long-tail prediction of \ac{DocRE} models trained on distant datasets: DocRED, UGDRE, DOREMI \textbf{(ours)}, and D+U (DOREMI for long-tail, UGDRE for frequent relations). Results are on the \textbf{Re-DocRED test set}; extreme long-tail refers to relations with $<100$ training instances. Best and second-best results are bolded and underlined.}

\label{tab:redocred-100}
\resizebox{\columnwidth}{!}{%
\begin{tabular}{ll|ccccc}
\toprule
Model & Distant Data & Precision & IgnPrec & Recall & F1 & Ign F1 \\ 
\midrule
\multirow{4}{*}{\begin{tabular}[c]{@{}l@{}}ATLOP \\ BERT\end{tabular}} & DocRED & 0.2532 & 0.1016 & 0.2977 & 0.2737 & 0.1515 \\
 & UGDRE & 0.3333 & 0.1940 & \textbf{0.4122} & \underline{0.3683} & 0.2639 \\
 & \cellcolor{ours} DOREMI & \cellcolor{ours} \textbf{0.5806} & \cellcolor{ours} \textbf{0.4902} & \cellcolor{ours} \underline{0.2748} & \cellcolor{ours} \textbf{0.3731} & \cellcolor{ours} \textbf{0.3522} \\
 & \cellcolor{ours} D+U & \cellcolor{ours} \underline{0.5161} & \cellcolor{ours} \underline{0.4208} & \cellcolor{ours} 0.2443 & \cellcolor{ours} 0.3316 & \cellcolor{ours} \underline{0.3097} \\  
\midrule
\multirow{4}{*}{\begin{tabular}[c]{@{}l@{}}ATLOP \\ RoBERTa\end{tabular}} & DocRED & 0.2711 & 0.1295 & 0.3435 & 0.3030 & 0.1881 \\
 & UGDRE & 0.3253 & 0.1765 & \textbf{0.4122} & \underline{0.3636} & 0.2471\\
 & \cellcolor{ours} DOREMI & \cellcolor{ours} \textbf{0.5962} & \cellcolor{ours} \textbf{0.5435} & \cellcolor{ours} 0.2366 & \cellcolor{ours} 0.3388 & \cellcolor{ours} \underline{0.3297} \\
 & \cellcolor{ours} D+U & \cellcolor{ours} \underline{0.4756} & \cellcolor{ours} \underline{0.3582} & \cellcolor{ours} \underline{0.2977} & \cellcolor{ours} \textbf{0.3662} & \cellcolor{ours} \textbf{0.3252} \\ 
\specialrule{1pt}{0pt}{0pt}
\multirow{4}{*}{\begin{tabular}[c]{@{}l@{}}DREEAM \\ BERT\end{tabular}} & DocRED & 0.3538 & 0.1923 & 0.1756 & 0.2347 & 0.1836 \\
 & UGDRE & 0.3929 & 0.2917 & \textbf{0.3359} & \underline{0.3621} &  \underline{0.3122} \\
 & \cellcolor{ours} DOREMI & \cellcolor{ours} \underline{0.4921} & \cellcolor{ours} \underline{0.4483} & \cellcolor{ours} 0.2366 & \cellcolor{ours} 0.3196 & \cellcolor{ours} 0.3098 \\
 & \cellcolor{ours} D+U & \cellcolor{ours} \textbf{0.5606} & \cellcolor{ours} \textbf{0.4821} & \cellcolor{ours} \underline{0.2824} & \cellcolor{ours} \textbf{0.3756} & \cellcolor{ours} \textbf{0.3562} \\ 
\midrule
\multirow{4}{*}{\begin{tabular}[c]{@{}l@{}}DREEAM \\ RoBERTa\end{tabular}} & DocRED & 0.3085 & 0.1667 & 0.2214 & 0.2578 & 0.1902 \\
 & UGDRE & 0.3966 & 0.3069 & \textbf{0.3511} & \textbf{0.3725} &  \underline{0.3276} \\
 & \cellcolor{ours} DOREMI & \cellcolor{ours} \textbf{0.5862} & \cellcolor{ours} \textbf{0.4783} & \cellcolor{ours} \underline{0.2595} & \cellcolor{ours} \underline{0.3598} & \cellcolor{ours} \textbf{0.3365} \\
 & \cellcolor{ours} D+U & \cellcolor{ours} \underline{0.5636} & \cellcolor{ours} \underline{0.4545} & \cellcolor{ours} 0.2366 & \cellcolor{ours} 0.3333 & \cellcolor{ours} 0.3112 \\  
\bottomrule
\end{tabular}%
}
\end{table}

\ac{DOREMI} proves particularly effective on the \emph{ignored} metrics, which evaluate only pairs unseen by the models during training.
Compared to UGDRE on long-tail prediction, DOREMI achieves a $+207.9\%$ relative increase in \textsc{ignPrecision} and a $+33.5\%$ gain in \textsc{ignF1}. ATLOP-BERT more than doubles UGDRE’s \textsc{ignPrecision} on unseen long-tail predicitions, while ATLOP-RoBERTa more than triples it.

\begin{figure*}[t]
    \centering
    \includegraphics[width=\textwidth]{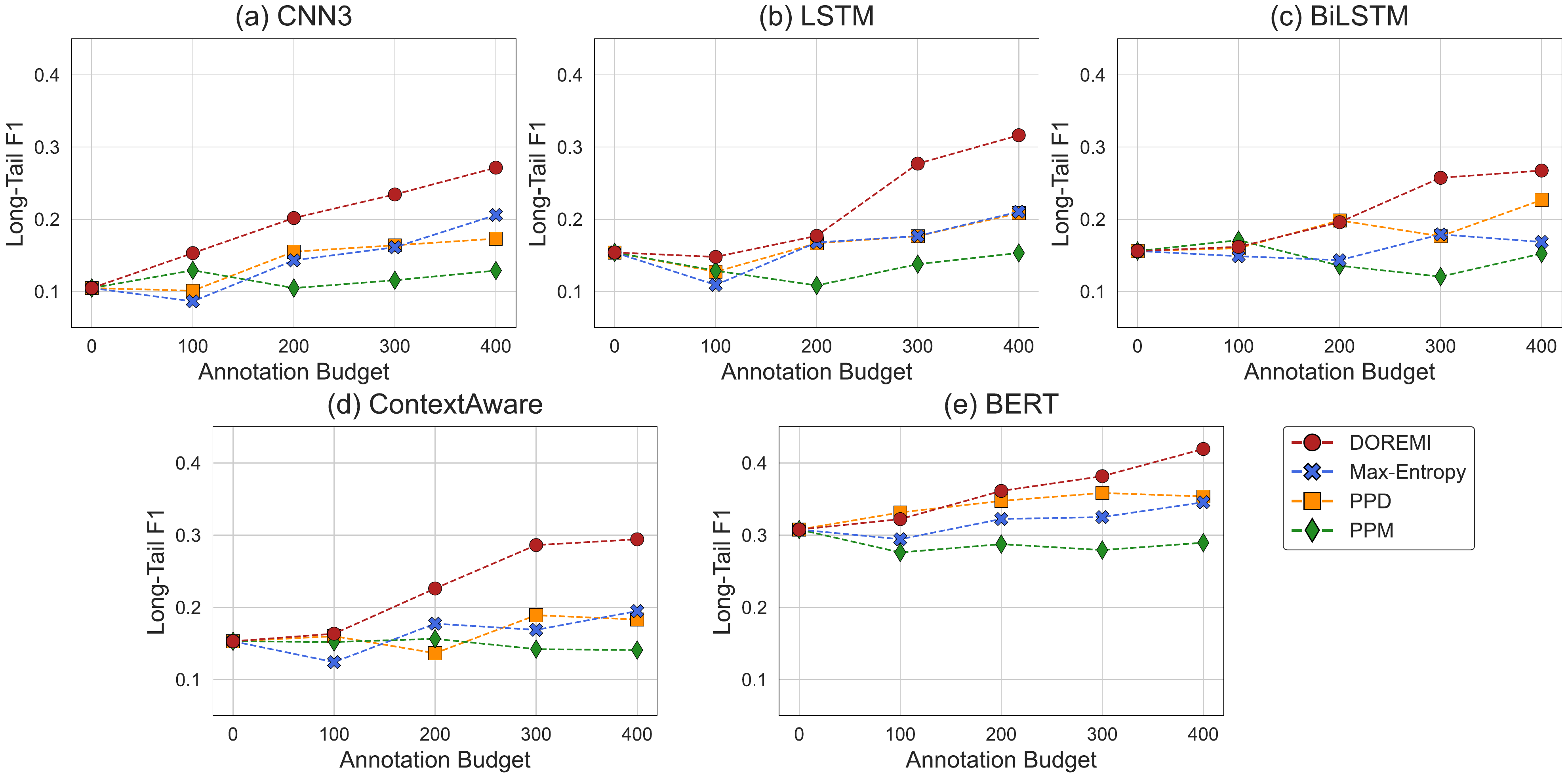}
    \caption{Micro-averaged F1 score of long-tail relations using different selection criteria computed on half of the DocRED dev dataset. Each subplot reports one of the five \ac{DocRE} core models exploited during iterative training by \ac{DOREMI}. Long-tail refers to relations with $< 100$ training instances. Annotation Budget "0" refers to the pre-training step, which is independent of the employed selection criteria.} 
    \label{fig:ablation}
\end{figure*}

\section{Discussion}
\label{sec:discussion}
This section investigates different strategies to select meaningful training examples to annotate, identifying the most suitable approach for our study (Subsection~\ref{subsec:sampling}). Subsection~\ref{subsec:sensitivity} describes the effect of setting the annotation budget $b=400$ for DocRED and motivates the infeasibility of a sensitivity analysis for \ac{DOREMI}.
In addition, we demonstrate the effectiveness of our disagreement-based sampling procedure by reporting the number of long-tail examples and N/A triples detected (Subsection~\ref{subsec:annotation}).

\subsection{Sampling Strategies}
\label{subsec:sampling}
We compare our disagreement-based strategy with \emph{Max-Entropy}, one of the most common sampling techniques~\cite{dagan_engelson-1995}. 
To establish the best solution to represent the disagreement between $n$ \ac{DocRE} models, we also consider a few alternatives to Equation~\ref{eqn:sc}.
We introduce a selection criterion based on the disagreement between relations called \emph{\ac{PPM}}. Instead of a probabilistic approach, \ac{PPM} considers the mean disagreement across all relations as a proxy to compute the disagreement between the models. In this scenario, the probability of disagreement on a given relation $r$ is computed as one minus the product of the probability that a given model predicts relation $r$ for the pair $(s,o)$, which we defined as $p_{\gamma_i}(r|(s,o))$ in Section \ref{subsec:disagreement}:
\begin{equation}
    \phi_\Gamma(r|(s,o)) = 1-\prod_{i=1}^n p_{\gamma_i}(r|(s,o))
\end{equation}
The sampling criterion selects the top-$k$ entity pairs with the highest mean probability of disagreement across all relations:
\begin{equation}
    \label{eqn:ppm}
    \arg \max_{(s,o)} \frac{\sum_{r \in R} \phi_\Gamma(r|(s,o))}{|R|}
\end{equation}

\ac{DOREMI} defines the probability of disagreement on a given relation $r$ using both the probability that a model predicts the relation and the probability that the model does not predict the relation (cfr. Section~\ref{subsec:disagreement}). Instead, in \emph{\ac{PPD}} the probability of disagreement only considers the probability that a model predicts the relation. In this case, the probability of disagreement on a given relation $r$ is defined as:
\begin{equation}
    \phi_\Gamma(r|(s,o)) = 1 - \prod_{i=1}^n p_{\gamma_i}(r|(s,o))
\end{equation}
Following the same steps as Section~\ref{subsec:disagreement}, the overall disagreement on the relation set $R$ for the entity pair $(s,o)$ is:
\begin{equation}
    \phi_\Gamma(s,o) = \prod_{r \in R} \phi_\Gamma(r|(s,o))
\end{equation}
We apply the same logarithmic transformation as Section~\ref{subsec:disagreement} and define the disagreement for the entity pair $(s,o)$ as:
\begin{equation}
    \psi_{\Gamma}(s,o) =  \sum_{r \in R} \log \phi_{\Gamma}(r|(s,o))
\end{equation}
Based on it, we define the sampling criterion as selecting the top‐$k$ entity pairs with the highest disagreement:
\begin{equation}
    \arg \max_{(s,o)} \psi_{\Gamma}(s, o)
\end{equation}

In \ac{DOREMI}, the training data set is iteratively enriched with annotated triples selected according to the chosen criterion. 
Thus, testing different selection criteria requires multiple annotation rounds for each criterion, which can be costly. 
Therefore, instead of sampling training examples from the DocRED \ac{DS} data and manually annotating the triples, we split the DocRED development dataset in half: one half serves for evaluation, while the other half is used as the sampling pool. We maintain the budget $b$ and $k$ used for the manual annotation, resulting in four iterations of \ac{DOREMI}. 

Figure~\ref{fig:ablation} reports the micro-averaged F1 score of long-tail relations at each iteration for the \ac{DocRE} core models using different selection criteria for the sampling module in \ac{DOREMI}. The F1 score is computed on half of the DocRED development dataset and considers long-tail relations with less than 100 training instances in the DocRED manual training dataset. 
The disagreement-based selection criterion exploited in \ac{DOREMI} shows a steady performance improvement across iterations for all core models. 
Between iterations $2$ and $5$, the steeper F1 improvement suggests more effective sampling. 
Other approaches yield minimal changes in the F1 score across iterations and, in some cases, result in performance degradation. Such a behaviour is evident in ContextAware and BERT when exploiting \ac{PPD} or in BiLSTM when using Max-Entropy. Notably, employing \ac{PPM} results in a lower F1 score than the pre-training step for ContextAware and BERT; hence, demonstrating it is not a good proxy to select training instances. 

\subsection{Sensitivity Analysis}
\label{subsec:sensitivity}
Figure~\ref{fig:ablation} reports \ac{DOREMI} micro-averaged F1 scores of long-tail relation computed on half of the DocRED dev dataset using different annotation budgets (\ac{DOREMI} is represented with red circled). All core models report a sizable improvement when increasing the number of annotations, motivating the need for the annotation budget for DocRED equal to $400$. We capped $b$ at $400$ to limit the amount of human annotation and because the improvement curve exhibits an inflection between $b=300$ and $b=400$, suggesting plateau performance gains beyond this point.

The sensitivity analysis for the other hyperparameters is infeasible for \ac{DOREMI}. Being a human-in-the-loop method, studying the effect of different values for each hyperparameter would require several human annotation rounds and a large annotation budget.

\subsection{Manual Annotation Analysis}
\label{subsec:annotation}
\begin{table}[t!]
\caption{Manual annotations statistics. For each dataset and each iteration, we report the number and the percentage over the total of annotated long-tail triples, frequent relations, N/A, and the total number of annotated triples. }
\label{tab:annotations-stats}
\centering
\footnotesize 
\begin{tabular}{l|cccc}
\toprule
Iteration & Long-Tail & Frequent & N/A & Triples \\ 
\midrule
\multicolumn{5}{l}{\textbf{(a) DOREMI iterative training with DocRED}} \\[0.5em]
Iteration 1 & 34 (34 \%) & 42 (42\%) & 24 (24\%) & 100 \\
Iteration 2 & 39 (39 \%) & 43 (43\%) & 18 (18\%) & 100\\
Iteration 3 & 49 (49 \%) & 36 (36\%) & 15 (15\%) & 100 \\
Iteration 4 & 60 (60 \%) & 14 (14\%) & 26 (26\%) & 100    \\
\midrule
Total & 182 (46 \%) & 135 (33\%) & 83 (21\%) & 400 \\ 
\specialrule{1pt}{0pt}{3pt}
\multicolumn{5}{l}{\textbf{(b) DOREMI iterative training with Re-DocRED}} \\[0.5em]
Iteration 1 & 172 (57 \%) & 122 (41\%) & 6 (2\%) & 300 \\
Iteration 2 & 178 (59 \%) & 99 (33\%) & 23 (8\%) & 300 \\
Iteration 3 & 183 (61 \%) & 84 (28\%) & 33 (11\%) &  300 \\
Iteration 4 & 206 (67 \%) & 50 (18\%) & 44 (15 \%) & 300 \\
\midrule
Total & 739 (62 \%) & 355 (29\%) & 106 (9\%) & 1,200 \\ 
\bottomrule
\end{tabular}%
\end{table}
This section reports some statistics on the manual annotation process. We investigate whether the sampling strategy employed in \ac{DOREMI} effectively identifies long-tail examples and the number of N/A triples detected. Table~\ref{tab:annotations-stats} reports the manual annotation statistics for both DocRED and Re-DocRED. In both datasets, the number of long-tail triples annotated grows at each iteration, demonstrating that models are getting better at predicting long-tail triples and the sampling strategy effectively samples them.
Notably, for the DocRED dataset, the growth is more pronounced. During the first iteration, the ratio of frequent to long-tail triples is nearly equal, whereas by the final iteration, long-tail relations account for 60\% of the annotated triples, while frequent triples make up only 14\%. 

The portion of N/A instances is marginal across all iterations and datasets. However, it grows at each iteration in Re-DocRED. A manual analysis of these cases revealed that the majority of N/A instances in Re-DocRED samples stem from incorrect entity annotations in the DocRED \ac{DS} dataset. This observation, coupled with the superior performance of models trained on Re-DocRED compared to those trained on DocRED, suggests that highly uncertain triples are often the result of such annotation errors. Nevertheless, incorporating N/A triples enhances model recall -- as demonstrated in Section~\ref{sec:results} -- by helping the model learn to distinguish between the presence and absence of a relation.

While the sampling strategy is designed to emphasize long-tail relations, it cannot eliminate the presence of frequent triples and N/A instances. However, including such examples contributes to a more balanced training signal, ultimately enabling the models trained with the \ac{DOREMI} denoised dataset to achieve strong performance. 

\section{Conclusions}
\label{sec:conclusion}

We introduced \ac{DOREMI}, a dataset enhancement framework that targets long-tail relation prediction while minimizing manual annotation. Leveraging iterative training, \ac{DOREMI} identifies Hard-To-Classify examples by measuring disagreement across multiple models.
Being optimized for long-tail predictions, \ac{DOREMI} can complement existing denoising approaches, such as UGDRE, which are better suited for frequent relations. 
The resulting denoised distantly supervised dataset can be used to train any off-the-shelf \ac{DocRE} model, yielding improved performance on long-tail relation prediction.
Experiments on DocRED and Re-DocRED show that \ac{DOREMI} significantly improves performance, especially in long-tail settings. 
Remarkably, annotating just $0.001\%$ of the DocRED \ac{DS} dataset yields precision improvements of up to a $+8.2\%$ overall and $+76.0\%$ in long-tail predictions over UGDRE -- the current state-of-the-art in label denoising -- as evaluated on the DocRED dev dataset. DOREMI also significantly boosts long-tail \textsc{ignPrecision} by $+137.6\%$, indicating better generalization to novel entity pairs unseen during training.
Similar findings hold for Re-DocRED, where labeling $0.003\%$ of the data leads to a $+16.2\%$ gain in long-tail precision and $+19.2\%$ on \textsc{ignPrecision}. On the \emph{extreme long-tail triples}, \ac{DOREMI} achieves a $+83.2\%$ relative increase in precision and $+207.9\%$ in \textsc{ignPrecision} compared to UGDRE.

These results highlight the effectiveness of disagreement-driven annotation 
-- enabling better generalization with negligible human effort.

\section*{Funding}
This work has received funding from the HEREDITARY Project as part of the European Union’s Horizon Europe research and innovation programme under grant agreement No GA 101137074. Views and opinions expressed are, however, those of the authors only and do not necessarily reflect those of the European Union. Neither the European Union nor the granting authority can be held responsible.

\section*{Contributions}
L.M. contributed to the design and development of DOREMI, creation of the distant denoised datasets, and manuscript preparation. 
S.M. contributed to the design and formalization of DOREMI and manuscript preparation.
G.S. contributed to the design and formalization of DOREMI, supervision and coordination of the teamwork, and manuscript revision.
All the authors contributed to the revision of the manuscript.

\section*{Competing Interests}
The authors declare no competing interests.

\section*{Declaration of Generative AI use}
During the preparation of this work, the authors used ChatGPT, Grammarly, and Comet as writing assistants, in particular to rephrase some parts of the writing. After using this tool/service, the authors reviewed and edited the content as needed and take full responsibility for the content of the published article.

\bibliographystyle{elsarticle-num-names}
\bibliography{custom}
\newpage
\appendix

\section{Macro-Averaged Results}
\label{app:macro-results}
\begin{table}[t!]
\caption{Macro-averaged performance of \ac{DocRE} models trained on distant datasets: DocRED, UGDRE, DOREMI \textbf{(ours)}, and D+U (DOREMI for long-tail, UGDRE for frequent relations). Results are on the \textbf{DocRED} dev set. Best and second-best results are bolded and underlined.}

\label{tab:results-macro-docred}
\resizebox{\columnwidth}{!}{%
\begin{tabular}{ll|ccccc}
\toprule
Model & Distant Data & Precision & IgnPrec &  Recall & F1 & Ign F1 \\ 
\midrule
\multirow{4}{*}{\begin{tabular}[c]{@{}l@{}}ATLOP \\ BERT\end{tabular}} & DocRED & 0.4499 & 0.2764 & 0.4456 & 0.4196 & 0.2917 \\ 
 & UGDRE & 0.4490 & 0.2951 & \textbf{0.5117} & \textbf{0.4572} & \underline{0.3358} \\ 
 & \cellcolor{ours} DOREMI & \cellcolor{ours} \textbf{0.4825} & \cellcolor{ours} \textbf{0.3693} & \cellcolor{ours} 0.4057 & \cellcolor{ours} 0.3997 & \cellcolor{ours} \textbf{0.3402} \\ 
 & \cellcolor{ours} D+U & \cellcolor{ours} \underline{0.4576} & \cellcolor{ours} \underline{0.3162} & \cellcolor{ours} \underline{0.4486} & \cellcolor{ours} \underline{0.4343} & \cellcolor{ours} 0.3351 \\ 
\midrule
\multirow{4}{*}{\begin{tabular}[c]{@{}l@{}}ATLOP \\ RoBERTa\end{tabular}}  & DocRED & 0.4497 & 0.2844 & \underline{0.4696} & 0.4299 & 0.3054 \\ 
 & UGDRE & 0.4497 & 0.3006 & \textbf{0.5481} & \textbf{0.4682} & 0.3495 \\ 
 & \cellcolor{ours} DOREMI & \cellcolor{ours} \underline{0.4502} & \cellcolor{ours} \underline{0.3433} & \cellcolor{ours} 0.4264 & \cellcolor{ours} 0.4128 & \cellcolor{ours} \underline{0.3505} \\ 
 & \cellcolor{ours} D+U & \cellcolor{ours} \textbf{0.4834} & \cellcolor{ours} \textbf{0.3479} & \cellcolor{ours} 0.4663 & \cellcolor{ours} \underline{0.4465} & \cellcolor{ours} \textbf{0.3541} \\
\specialrule{1pt}{0pt}{0pt}
\multirow{4}{*}{\begin{tabular}[c]{@{}l@{}}DREEAM \\ BERT\end{tabular}} & DocRED & 0.4711 & 0.3277 & 0.3310 & 0.3566 & 0.2764 \\ 
 & UGDRE & \underline{0.4801} & 0.3408 & \underline{0.4255} & \underline{0.4279} & 0.3420 \\ 
 & \cellcolor{ours} DOREMI & \cellcolor{ours} 0.4444 & \cellcolor{ours} \underline{0.3430} & \cellcolor{ours} \textbf{0.4519} & \cellcolor{ours} 0.4260 & \cellcolor{ours} \underline{0.3633} \\ 
 & \cellcolor{ours} D+U & \cellcolor{ours} \textbf{0.5199} & \cellcolor{ours} \textbf{0.3916} & \cellcolor{ours} 0.4172 & \cellcolor{ours} \textbf{0.4415} & \cellcolor{ours} \textbf{0.3673} \\ 
\midrule
\multirow{4}{*}{\begin{tabular}[c]{@{}l@{}}DREEAM \\ RoBERTa\end{tabular}} & DocRED & 0.4644 & 0.3200 & 0.3709 & 0.3851 & 0.2997 \\ 
 & UGDRE & \underline{0.4811} & 0.3364 & \textbf{0.4626} & \underline{0.4477} & 0.3543 \\ 
 & \cellcolor{ours} DOREMI & \cellcolor{ours} 0.4749 & \cellcolor{ours} \underline{0.3742} & \cellcolor{ours} \underline{0.4496} & \cellcolor{ours} 0.4312 & \cellcolor{ours} \underline{0.3673} \\ 
 & \cellcolor{ours} D+U & \cellcolor{ours} \textbf{0.5325} & \cellcolor{ours} \textbf{0.3993} & \cellcolor{ours} 0.4416 & \cellcolor{ours} \textbf{0.4542} & \cellcolor{ours} \textbf{0.3741} \\ 
\bottomrule
\end{tabular}%
}
\end{table}

\begin{table}[t!]
\caption{Macro-averaged performance of \ac{DocRE} models trained on distant datasets: DocRED, UGDRE, DOREMI \textbf{(ours)}, and D+U (DOREMI for long-tail, UGDRE for frequent relations). Results are on the \textbf{Re-DocRED} test set. Best and second-best results are bolded and underlined.}

\label{tab:results-macro-redocred}
\resizebox{\columnwidth}{!}{%
\begin{tabular}{ll|ccccc}
\toprule
Model & Distant Data & Precision & IgnPrec &  Recall & F1 & Ign F1 \\ 
\midrule
\multirow{4}{*}{\begin{tabular}[c]{@{}l@{}}ATLOP \\ BERT\end{tabular}} & DocRED & 0.6614 & 0.5093 & 0.3016 & 0.3698 & 0.3075 \\ 
 & UGDRE & \underline{0.7088} & \underline{0.6189} & \textbf{0.5970} & \textbf{0.6264} & \textbf{0.5702} \\ 
 & \cellcolor{ours} DOREMI & \cellcolor{ours} 0.6922 & \cellcolor{ours} 0.5801 & \cellcolor{ours} 0.5354 & \cellcolor{ours} 0.5781 & \cellcolor{ours} 0.5181 \\ 
 & \cellcolor{ours} D+U & \cellcolor{ours} \textbf{0.7388} & \cellcolor{ours} \textbf{0.6578} & \cellcolor{ours} \underline{0.5490} & \cellcolor{ours} \underline{0.6107} & \cellcolor{ours} \underline{0.5641} \\ 
\midrule
\multirow{4}{*}{\begin{tabular}[c]{@{}l@{}}ATLOP \\ RoBERTa\end{tabular}} & DocRED & 0.6691 & 0.5218 & 0.3354 & 0.3983 & 0.3393 \\ 
 & UGDRE & \underline{0.7064} & \underline{0.6195} & \textbf{0.6075} & \textbf{0.6358}& \textbf{0.5774} \\ 
 & \cellcolor{ours} DOREMI & \cellcolor{ours} 0.6951 & \cellcolor{ours} 0.5885 & \cellcolor{ours} 0.5275 & \cellcolor{ours} 0.5699 & \cellcolor{ours} 0.5150 \\ 
 & \cellcolor{ours} D+U & \cellcolor{ours} \textbf{0.7302} & \cellcolor{ours} \textbf{0.6500} & \cellcolor{ours} \underline{0.5760} & \cellcolor{ours} \underline{0.6245} & \cellcolor{ours} \underline{0.5760} \\ 
\specialrule{1pt}{0pt}{0pt}
\multirow{4}{*}{\begin{tabular}[c]{@{}l@{}}DREEAM \\ BERT\end{tabular}} & DocRED & 0.7548 & 0.6040 & 0.2461 & 0.3236 & 0.2826 \\ 
 & UGDRE & \underline{0.7238} & \underline{0.6528} & \textbf{0.5813} & \textbf{0.6253} & \textbf{0.5843} \\ 
 & \cellcolor{ours} DOREMI & \cellcolor{ours} 0.7100 & \cellcolor{ours} 0.6110 & \cellcolor{ours} 0.5400 & \cellcolor{ours} 0.5918 & \cellcolor{ours} 0.5387 \\ 
 & \cellcolor{ours} D+U & \cellcolor{ours} \textbf{0.7628} & \cellcolor{ours} \textbf{0.6990} & \cellcolor{ours} \underline{0.5232} & \cellcolor{ours} \underline{0.6013} & \cellcolor{ours} \underline{0.5649} \\ 
\midrule
\multirow{4}{*}{\begin{tabular}[c]{@{}l@{}}DREEAM \\ RoBERTa\end{tabular}}& DocRED & \underline{0.7642} & 0.6453 & 0.2834 & 0.3633 & 0.3245 \\ 
 & UGDRE & 0.7597 & \underline{0.6929} &  \textbf{0.6164} & \textbf{0.6599} & \textbf{0.6216} \\ 
 & \cellcolor{ours} DOREMI & \cellcolor{ours} 0.7605 & \cellcolor{ours} 0.6591 & \cellcolor{ours} \underline{0.5661} & \cellcolor{ours} 0.6260 & \cellcolor{ours} 0.5725 \\ 
 & \cellcolor{ours} D+U & \cellcolor{ours} \textbf{0.7963} & \cellcolor{ours} \textbf{0.7311} & \cellcolor{ours} 0.5522 & \cellcolor{ours} \underline{0.6306} & \cellcolor{ours} \underline{0.5983} \\  
\bottomrule
\end{tabular}%
}
\end{table}

Tables~\ref{tab:results-macro-docred} and~\ref{tab:results-macro-redocred} report the macro-average performance of ATLOP and DREEAM trained on different datasets and tested on DocRED and Re-DocRED, respectively. Macro-averaging computes the metric independently for each class and then takes the average, treating all classes equally regardless of their frequency. This method is particularly useful for assessing performance on rare or underrepresented classes. Thus, in our case, it makes sense to display the macro-average performance on all relations. 

In both datasets, the macro-averaged performance confirms our findings in Section~\ref{sec:results}, demonstrating how \ac{DOREMI} alone or combined with UGDRE improves the \emph{ignored} metrics -- \textsc{ignPrecision} and \textsc{ignF1} --, aiding models to generalize to unseen pairs. In particular, when \ac{DOREMI} exploits DocRED for iterative training (Table~\ref{tab:results-macro-docred}), \ac{DOREMI} shows an improvement of up to $+25.1\%$ in \textsc{ignPrecision} and $+6.2\%$ in \textsc{ignF1} compared to UGDRE. 

When \ac{DOREMI} exploits Re-DocRED for iterative training (Table~\ref{tab:results-macro-redocred}), combining \ac{DOREMI} with UGDRE (dataset D+U) consistently outperform UGDRE in terms of precision and \textsc{ignPrecision}, reporting a relative of up to $+5.4\%$ and $+7.1\%$, respectively.

\section{Large Language Models Performance}
\label{app:llms}
\begin{table*}[t!]
\caption{Performance of \ac{DocRE} models trained on the DOREMI distant dataset and \acp{LLM}. Best results are bolded.}

\label{tab:results-llms}
\resizebox{\textwidth}{!}{%
\begin{tabular}{ll|ccccc|ccccc}
\toprule
\multirow{2}{*}{Category} & \multirow{2}{*}{Model} & \multicolumn{5}{c|}{Full Dataset} & \multicolumn{5}{c}{Long-Tail Triples} \\ 
\multicolumn{1}{l}{} & & Precision & Ign Prec &  Recall & F1 & Ign F1 & Precision & Ign Prec & Recall & F1 & Ign F1 \\ 
\midrule
\multicolumn{12}{l}{\textbf{(a) Evaluated on DocRED dev dataset (long-tail relations with <100 training instances)}} \\[0.5em]
\multirow{2}{*}{LLMs} & GPT-4.1 & 0.1986 & - - - & 0.0528 & 0.0834 & - - - & 0.0654 & - - - & 0.0470 & 0.0547 & - - - \\
& Llama-3.3-70B & 0.0236 & - - - & 0.0025 & 0.0046 & - - - & 0.000 & - - - & 0.0000 & 0.0000 & - - - \\
\midrule
\multirow{2}{*}{BERT} & ATLOP-DOREMI & 0.5722 & 0.4384 & 0.6160 & 0.5933 & 0.5123 & 0.3634 & 0.2701 & 0.1946 & 0.2535 & 0.2262 \\ 
& DREEAM-DOREMI & 0.5765 & 0.4433 & 0.6341 & 0.6039 & 0.5218 & 0.3928 & 0.3058 & 0.2483 & 0.3043 & 0.2741 \\
\midrule
\multirow{2}{*}{RoBERTa} & ATLOP-DOREMI & 0.5788 & 0.4453 & 0.6242 & 0.6006 & 0.5198 & 0.4200 & 0.3324 & 0.2362 & 0.3024 & 0.2762 \\ 
& DREEAM-DOREMI & \textbf{0.5904} & \textbf{0.4564} & \textbf{0.6513} & \textbf{0.6194} & \textbf{0.5367} & \textbf{0.4332} & \textbf{0.3492} & \textbf{0.2523} & \textbf{0.3189} & \textbf{0.2930} \\
\midrule
\multicolumn{12}{l}{\textbf{(b) Evaluated on Re-DocRED test dataset (long-tail relations with <300 training instances)}} \\[0.5em]
\multirow{2}{*}{LLMs} & GPT-4.1 & 0.1612 & - - - & 0.0301 & 0.0579 & - - - & 0.0093 & - - - & 0.0153 & 0.0116 & - - - \\
& Llama-3.3-70B & 0.0198 & - - - & 0.0015 & 0.0028 & - - - & 0.000 & - - - & 0.0000 & 0.0000 & - - - \\
\midrule
\multirow{2}{*}{BERT} & ATLOP-DOREMI & 0.7480 & 0.6297 & 0.6914 & 0.7186 & 0.6591 & 0.6023 & 0.5004 & 0.3628 & 0.4299 & 0.3795 \\ 
& DREEAM-DOREMI & 0.7701 & 0.6576 & 0.7037 & 0.7354 & 0.6799 & 0.7250 & 0.6640 & 0.4210 & 0.5326 & 0.5153 \\
\midrule
\multirow{2}{*}{RoBERTa} & ATLOP-DOREMI & 0.7486 & 0.6311 & 0.7023 & 0.7247 & 0.6648 & 0.6232 & 0.5315 & 0.3489 & 0.4206 & 0.3784 \\ 
& DREEAM-DOREMI & \textbf{0.7939} & \textbf{0.6894} & \textbf{0.7267} & \textbf{0.7588} & \textbf{0.7075} & \textbf{0.7575} & \textbf{0.7032} & \textbf{0.4498} & \textbf{0.5644} & \textbf{0.5486} \\
\bottomrule
\end{tabular}%
}
\end{table*}

Recent works explored \acp{LLM} for \ac{DocRE}~\cite{zhang_etal-2025a,zhang_etal-2025b,xue_etal-2024}. To enhance their performance, logical reasoning~\cite{zhang_etal-2024,qi_etal-2024} and data augmentation~\cite{fan_etal-2024,gao_etal-2024,modarressi_etal-2024} have been proposed to integrate additional filtering. Although this is a promising avenue, \acp{LLM} still proves to be ineffective for \ac{DocRE}. 
The primary reason is that \acp{LLM} often overestimate positive relations between entities, as they struggle to handle the large proportion of negative examples present in \ac{DocRE} datasets. A secondary factor is their difficulty with multi-label classification. Indeed, \acp{LLM} frequently predict a single relation even when multiple relations are expressed for the same entity pair within a document~\cite{zhang_etal-2025a}.

\begin{figure*}[t!]
    \centering
    \includegraphics[width=\textwidth]{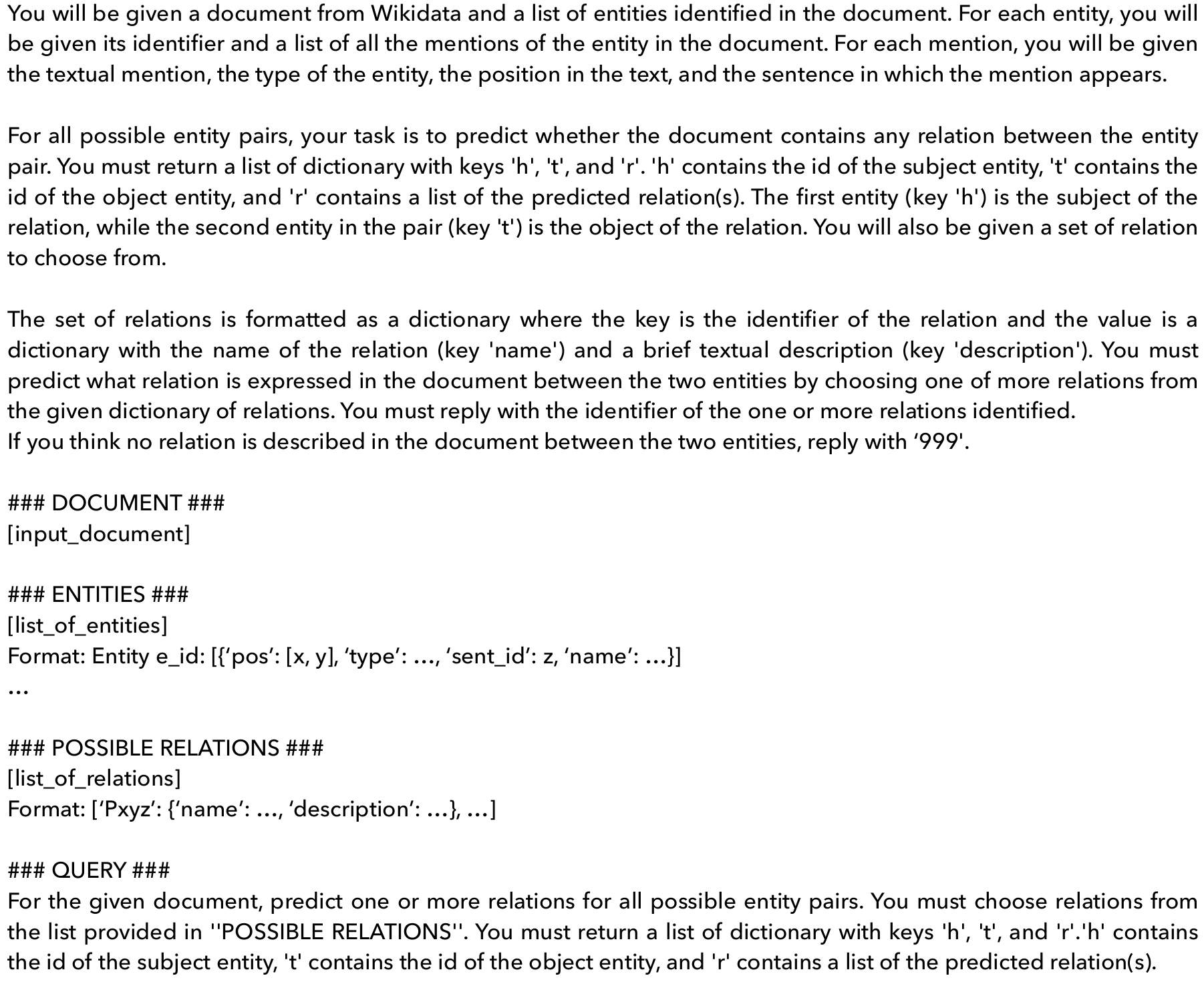}
    \caption{Prompt template for \ac{DocRE}. For each document, information between square brackets was integrated accordingly. The same prompt is used for both \acp{LLM}.}
    \label{fig:prompt}
\end{figure*}

This section compares the performance of models trained with DOREMI \acp{DDS} and \acp{LLM}. To obtain the \acp{LLM} predictions, inspired by~\cite{fan_etal-2024}, we exploit a zero-shot prompt-based approach
using the same prompt for all LLMs. We use two \acp{LLM}: Llama 3.3 70B Instruct~\cite{grattafiori2024llama3herdmodels} and GPT-4.1~\cite{openai2024gpt4technicalreport}. 
These two models were chosen primarily for their availability and compatibility with our computational resources.
To run GPT-4.1, we used the Azure OpenAI Batch API in the Azure AI Foundry Portal and set the LLM temperature to $0.75$. To run Llama 3.3 70B Instruct, we developed a Python script that exploit Ollama.\footnote{\url{https://github.com/ollama/ollama}} We empirically set \texttt{num\_predict} to $1,000$, \texttt{temperature} to $0.7$, and \texttt{top\_p} to $0.1$. Due to limited computational resources, we opted for \texttt{llama3.3:70b-instruct-q4\_0},~\footnote{\url{https://ollama.com/library/llama3.3:70b-instruct-q4_0}} a 4-bit quantized version of Llama 3.3 70B Instruct.
Llama 3.3 70B Instruct was run on a Mac Studio 2025 with 512 GB of memory and the Apple M3 Ultra chip. 
We report the prompt template we used in Figure~\ref{fig:prompt}. 

Table~\ref{tab:results-llms} reports the micro-averaged results for the \acp{LLM} compared with two state-of-the-art models trained with the \ac{DOREMI} \ac{DDS}. Table~\ref{tab:results-llms} (a) reports the performance of all the models evaluated in the DocRED dev dataset. ATLOP and DREEAM are trained on the \ac{DOREMI} \ac{DDS} produced using the DocRED dataset during training. Table~\ref{tab:results-llms} (b) evaluates the models in the Re-DocRED test dataset. In this configuration, ATLOP and DREEAM are trained on the \ac{DOREMI} \ac{DDS} constructed with iterative training on the Re-DocRED dataset during training.

Both \acp{LLM} underperform relative to state-of-the-art models, exhibiting a substantial drop in performance. GPT outperforms LLama, achieving a \textsc{precision} of nearly $20\%$ and an \textsc{F1} score of $8\%$ on the DocRED development set, while on the Re-DocRED test dataset \textsc{precision} reaches $16\%$ and \textsc{F1} is approaching $6\%$.
GPT struggles in \textsc{recall}, obtaining only $5\%$ on the DocRED development set.
Indeed, GPT predicts $3,422$ triples for the documents in the DocRED development dataset -- regardless of correctness -- while the ground truth contains $12,275$ triples. Llama confirms this pattern, with \textsc{recall} being ten times lower than \textsc{precision}. In this case, the \ac{LLM} returned only $1,351$ triples in the DocRED development documents, irrespective of whether they were correct.
Furthermore, GPT and Llama predicted $11$ and $3$ relations, respectively, outside of DocRED relations, highlighting the tendency of \acp{LLM} to hallucinate.

Results indicate that \acp{LLM} remain less effective than current approaches for \ac{DocRE}. Furthermore, these findings justify our decision not to use \acp{LLM} to annotate new data for inclusion in the DOREMI loop, as their predictions are unreliable and may introduce even more noise than \ac{DS}.

\end{document}